\documentclass{article}

\usepackage[numbers]{natbib}

\usepackage[preprint]{neurips_2023}

\usepackage{amsmath, amssymb, amsthm, amsbsy, amscd, bm, bbm,mathrsfs}

 \usepackage{graphicx}

\newcommand{\g}{\mathbf{g}}

\newcommand{\I}{\mathbf{I}}

\newcommand{\w}{\mathbf{w}}

\newcommand{\Z}{\mathbf{Z}}

\newcommand{\E}{\mathbb{E}}

\newcommand{\cN}{\mathcal{N}}

\newcommand{\cL}{\mathcal{L}}
\newcommand{\clip}{\mathcal{C}}

\usepackage{microtype}
\usepackage{graphicx}
\usepackage{subfigure}
\usepackage{multirow}
\usepackage{multicol}
\usepackage{graphicx}
\usepackage{caption}
\usepackage{listings}

\theoremstyle{plain}
\newtheorem{theorem}{Theorem}[section]

\newtheorem{lemma}{Lemma}[section] %
\newtheorem{corollary}{Corollary}[theorem] %
\theoremstyle{definition}
\newtheorem{definition}[theorem]{Definition}
\newtheorem{assumption}[theorem]{Assumption}
\theoremstyle{remark}
\newtheorem{remark}[theorem]{Remark}

\usepackage[utf8]{inputenc} %
\usepackage[T1]{fontenc}    %
\usepackage{hyperref}       %
\usepackage[capitalize,noabbrev]{cleveref}
\usepackage{url}            %
\usepackage{booktabs}       %
\usepackage{amsfonts}       %
\usepackage{nicefrac}       %
\usepackage{microtype}      %
\usepackage{xcolor}         %
\usepackage{comment}

\crefname{lemma}{Lemma}{Lemmas}

\title{Coupling public and private gradient provably helps optimization\thanks{This work has been submitted to the IEEE for possible publication. Copyright may be transferred without notice, after which this version may no longer be accessible.}}

\author{%
  Ruixuan Liu\thanks{work done during internship} \\
  Emory University\\
  \texttt{ruixuan.liu2@emory.edu} \\
  \And
  Zhiqi Bu \\
  AWS AI \\
  \texttt{zhiqibu@amazon.com} \\
  \And
  Yu-xiang Wang \\
  University of California, Santa Barbara \\
  \texttt{yuxiangw@cs.ucsb.edu} \\
  \And
  Sheng Zha \\
  AWS AI \\
  \texttt{zhasheng@amazon.com} \\
  \And
  George Karypis \\
  AWS AI \\
  \texttt{gkarypis@amazon.com}
}

\begin{document}
\maketitle
\begin{abstract}
The success of large neural networks is crucially determined by the availability of data. 
It has been observed that training only on a small amount of public data, or privately on the abundant private data can lead to undesirable degradation of accuracy.
In this work, we leverage both private and public data to improve the optimization, by coupling their gradients via a weighted linear combination.
We formulate an optimal solution for the optimal weight in the convex setting to indicate that the weighting coefficient should be hyperparameter-dependent.
Then, we prove the acceleration in the convergence of non-convex loss and the effects of hyper-parameters such as privacy budget, number of iterations, batch size, and model size on the choice of the weighting coefficient. 
We support our analysis with empirical experiments across language and vision benchmarks, and provide a guideline for choosing the optimal weight of the gradient coupling.
\end{abstract}

\section{Introduction}
Nowadays, the superior learning performance achieved by deep neural networks is backed up by the availability of a large amount of representative data.
However, only a small amount of public data can be free for model training due to privacy concerns, especially in sensitive fields such as health care and finance.
For instance, some datasets cannot be publicly available due to proprietary policies, e.g. Google's JFT datasets (containing 300 million to 3 billion images) that serve as the training data of state-of-the-art large vision models~\cite{riquelme2021scaling}.
In addition, General Data Protection Regulation (GDPR)~\cite{EUdataregulations2018} requires that the private information, such as medical data in hospitals, should be processed in a manner that ensures appropriate confidentiality for potential attacks.

Above privacy concerns are raised by existing privacy risks.
Recent works have shown that if sensitive data are utilized as plaintext for model training, the privacy attackers may launch attacks to infer whether an individual's data exist in the training dataset~\cite{shokri2017membership}, or reconstruct the sensitive information in training data (e.g., password, address) \cite{carlini2021extracting}, or even recover the raw data \cite{zhu2019deep,haim2022reconstructing} via the machine learning service API or the published model parameters.

To limit the possible information leakage of any single training sample, differentially private (DP) deep learning~\cite{abadi2016deep} emerges as a solution by clipping each per-sample gradient and injecting random noise in the optimization. 
While DP optimization has achieved exciting results on large models including ViT~\cite{mehta2022large,bu2022scalable}, RoBERTa~\cite{li2021large,bu2022automatic}, and GPT~\cite{yu2021differentially,li2021large}, it is commonly observed that some accuracy degradation is in place compared to the standard non-DP optimization. 
For example, GPT-2-Medium has a 5.1 drop of BLEU score with $(6.8, 1e-5)$-DP compared to the performance of a non-DP model~\cite{yu2021differentially}.
A stronger privacy guarantee will aggravate the utility drop, which is not acceptable by high-stakes tasks.

\begin{table}[!htb]
    \centering
    \caption{Comparison of optimization with public and private data.}
    \label{tab:compare}
    \begin{tabular}{c|c|c|c}
    \toprule[1pt]
         & dataset & accuracy & DP guarantee \\
    \midrule[1pt]
        only public & small & low & -- \\
        only private & large & low & yes \\
        public \& private & large & high & yes \\
    \bottomrule[1pt]
    \end{tabular}
\end{table}
Therefore, to maximize the data efficiency (which is limited if we only have access to the public data) and to overcome the optimization challenge of DP training (which is difficult if we only utilize the private data), it is crucial to combine the public and private data under the DP framework, as shown in \Cref{tab:compare}.
Specifically, recent literature has proposed multiple strategies when only a small amount of in-distribution data are public but the private data are abundant \cite{golatkar2022mixed, zhou2020bypassing, kairouz2020fast, abadi2016deep}, with the possible availability of out-of-distribution public data for pretraining.
Along one line of works improve the performance of DP training in an implicit way by applying the statistical information of public per-sample norm~\cite{andrew2021differentially,golatkar2022mixed}, the sub-space structure~\cite{kairouz2020fast, zhou2020bypassing} or the second moments of the public samples~\cite{li2022private} into the DP optimization.

Another line of works~\cite{golatkar2022mixed,amid2022public,ferrando2021combining} indicates a more explicit way of utilizing public data via merging the public gradient together with the private gradient through a linear combination with the weighting coefficient $\alpha$, which will be introduced in \Cref{sec:linear combine}. 
For example, AdaMix~\cite{golatkar2022mixed} linearly combines the public and private gradient after applying the adaptive clipping and the adaptive projection.
DPMD~\cite{amid2022public} takes the private gradient as the linear term and regularizes with the public gradient as a mirror map. 
Since the computational cost is too high to apply in practice, DPMD is approximated to a linear combination of public and private gradients in all their experiments.

Our work analyzes the linear combination of public and private gradients in depth. 
To mark the difference between our method and previous ones~\cite{amid2022public, golatkar2022mixed}, firstly, we note that previous works choose the weighting coefficient $\alpha$ in an ad-hoc way as a constant. 
In sharp contrast, from the perspective of model convergence, our analysis on both convex and non-convex loss shows that the optimal $\alpha$ should be determined by all training hyperparameters, and thus improves over previous methods.
Secondly, though DPMD and AdaMix show an improvement of mixed training over DP training, the advantage of mixed training compared with training with only public data is still unknown.
We are motivated to analyze such an advantage through the lens of the weighting coefficient $\alpha$.

Our contributions can be summarized as follows:
1) We analyze the convergence of differentially private optimization that uses a linear combination of gradients from the in-distribution public and private data.
2) We show that the optimal weighting coefficients $\alpha$ between the public gradient and the private gradient are determined by multiple hyperparameters, including the privacy budget, number of iterations, batch size and model size.
In addition, the guideline for choosing $\alpha$ is compatible with existing mixed training methods and general optimizers.
3) We establish benchmarks on popular datasets and empirically demonstrate that coupling public and private gradients with a well-chosen $\alpha$ is effective for merging the gap between mixed training and the upper bound of non-DP training over the whole dataset.

\section{Preliminaries}
In mixed training, we have a public dataset $D_\textup{pub} := \{ z_1, \cdots, z_{n_\textup{pub}} \} $
and a private dataset $D_\textup{priv} := \{z_1, \cdots, z_{n_\textup{priv}}\}$, where $z=(x, y) \in \mathcal{D}$ is a data sample with feature and label.
With model parameters $\w \in \mathcal{W}$, we define $l: \mathcal{D} \times \mathcal{W} \rightarrow \mathbb{R}$ as the loss function.
$l(z)$ denotes the loss value given a sample $z\sim \mathcal{D}$ and the gradient for a sample $z_i$ is $\g_i=\bigtriangledown l_i(\w)$.

\subsection{Differential Privacy}
We follow the standard $(\epsilon, \delta)$-DP to measure the privacy risk, and smaller $\epsilon$ and $\delta$ indicate a stronger privacy guarantee. Given two neighboring datasets $\mathcal{S}, \mathcal{S}^\prime$ that differ by one sample,
Definition \ref{def-dp} bounds the worst-case information leakage so that the output difference of two datasets is indistinguishable.
Notice that, we only need $(\epsilon, \delta)$-DP for the private dataset, and the privacy budget for the public dataset is $(\infty, 1)$-DP.
\begin{definition}[Differential Privacy~\cite{dwork2014algorithmic}]\label{def-dp}
A randomized algorithm $M$ is ($\epsilon, \delta$)-differentially private if for any two neighbouring datasets $(\mathcal{S},\mathcal{S}^\prime)$, and any output event $E$:
\begin{align}
    \mathbb{P}[M(\mathcal{S}) \in E] \le e^\epsilon \mathbb{P}[M(\mathcal{S}^\prime) \in E] + \delta.
\end{align}
\end{definition}

Without loss of generality, we leverage
$\mu$-GDP \cite{dong2022gaussian} as an analytical tool, which indicates that asymptotically ($T\to \infty$), 
\begin{align}
\epsilon = \mu^2 + \mu\sqrt{2\log(1/\delta)} \text{ and }\mu = \frac{B}{n}\sqrt{T(e^{1/\sigma^2}-1)}
\label{eq:mu GDP}
\end{align}

Instead of studying two hyperparamters $(\epsilon,\delta)$, we can focus on $\mu$, which is monotone and unique to $(\epsilon,\delta)$.
Alternatively, similar conclusion can be drawn for $\rho$-tCDP~\cite{bun2018composable} by replacing $\mu^2$ with $\rho$.

\subsection{Differentially Private Optimization}
\label{sec:dp opt}
For public data, optimizers such as SGD and Adam update $\w$ with the public gradient in \eqref{eq:pub_grad}.
For private data, private gradient in \eqref{eq:priv_grad} is applied for DP guarantee with the per-sample gradient clipping function $C$ and the noise perturbation $\sigma R\cdot\mathcal{N}(0,\I)$.
\begin{align}
\text{Public gradient: }& \sum_j\g_j 
\label{eq:pub_grad}
\\
\text{Private gradient: }&  \sum_i \g_i \cdot C(\|\g_i\|; R) + \sigma R\cdot \mathcal{N}(0, \I).\label{eq:priv_grad}
\end{align}

Specifically, the noise magnitude $\sigma$ can be derived by the privacy accounting theory \cite{dwork2014algorithmic,abadi2016deep} for a given privacy budget $(\epsilon, \delta)$.
The threshold for the gradient clipping $R$ ensures $C(\|\g_i\|;R)\leq R/\|\g_i\|$, i.e. the gradient norm after clipping is at most $R$. 
While there exist a number of clipping functions, we apply the automatic clipping \cite{bu2022automatic} with $C(\|\g_i\|)=\frac{1}{\|\g_i\|+1}$ by setting $R=1$ and $\gamma=1$ because it eliminates the need to tune $R$, and it has been empirically verified to achieve state-of-the-art accuracy across computer vision and language tasks.
For a fair comparison, we apply the same clipping function for all other mixed training methods in our experiments.

\section{Linearly combining public and private gradients}
\label{sec:linear combine}
We follow a practical and classic setting~\cite{golatkar2022mixed, amid2022public, ferrando2021combining, jorgensen2015conservative} where there is a public dataset of size $n_\textup{pub}$ and a private dataset of size $n_\textup{priv}\gg n_\textup{pub}$, both following the same distribution. 
Denoting the Gaussian noise $\Z \sim \cN(0, \textbf{I})$ and the model parameters $\w$, 
one can mix the in-distribution public and private gradients in training via a linear combination as:
\begin{align*}
\w_{t+1}-\w_t
&=
-\left[ \eta_\textup{pub}\sum_{j}^{B} \g_{t, j} + 
\eta_\textup{priv} \left(\sum_{i}^{B} C_{t,i}\g_{t, i}+ \sigma R \cdot \Z \right)\right]
\end{align*}
Here $B$ is the batch size, $\eta_\text{pub}$(w.r.t. $\eta_\text{priv}$) is the learning rate for the public (w.r.t. private) gradient, and $\g_{t, i}$ is the per-sample gradient for sample $i$ at the $t^\text{th}$ iteration.
$C_{t, i}$ is the per-sample clipping factor, $\sigma$ is the noise multiplier that determines the privacy risk, and $R$ is the clipping threshold. 
By setting $\eta_\textup{pub}=\eta\alpha, \eta_\textup{priv}=\eta(1-\alpha)$ and $0 \le \alpha \le 1$, we can equivalently view the updating as:
\begin{align*}
\w_{t+1}-\w_t = -\eta_t\left[
\alpha\sum_{j}^{B} \g_{t, j} +(1-\alpha)\cdot \left( \sum_{i}^{B} C_{t,i}\g_{t, i}+ \sigma R\cdot \Z \right)
\right].
\end{align*}

As shown in \Cref{tab:compare linear combination}, such DP-SGD/Adam has previously been adopted under different choices of $\alpha$, $R$ and $C$. For instance, DPMD \cite{amid2022public} empirically uses $\alpha_t=1-\cos\frac{\pi t}{2K}$, where $K$ is the hyperparameter to control the rate of change in $\alpha_t$, so as to mimic the cosine annealing of learning rate; \cite{ferrando2021combining,jorgensen2015conservative} use a static choice of $\alpha=\frac{n_\textup{pub}}{n_\textup{pub}+n_\textup{priv}}$ (very small if $n_\textup{pub} \ll n_\textup{priv}$), though the setting is different and the per-sample gradient clipping is not used; AdaMix \cite{golatkar2022mixed} uses the adaptive clipping threshold $R_t$ as the 90\% quantile of per-sample gradient norms, as well as a simple static $\alpha=0.5$\footnote{Notice that AdaMix uses $\alpha=0.5$ when the objective the sum of per-sample losses and it becomes $\frac{n_\textup{pub}}{n_\textup{pub}+n_\textup{priv}}$ when the objective is the mean of losses. In both cases, the choice of $\alpha$ does not take $B,d,T,\epsilon$ into consideration.}.

\begin{table}[t]
\caption{Comparison to other mixed training methods.}
\label{tab:compare linear combination}
\centering
\begin{tabular}{c|c|c}
\toprule[1pt]
Methods & $\alpha$ & $C(\|\g_i\|)$ \\
\midrule[1pt]
Coupling (ours)& $\alpha(B,d,n,T,\epsilon)$ & $\frac{1}{\|\g_i\|+1}$ \\
DPMD~\cite{amid2022public} & 1-$\cos\frac{\pi t}{2K}$& 1\\
AdaMix~\cite{golatkar2022mixed} &0.5 & $\min\{\frac{R}{\|\g_i\|},1\}$\\
SampleMechanism~\cite{ferrando2021combining,jorgensen2015conservative} & $\frac{n_\textup{pub}}{n_\textup{pub}+n_\textup{priv}}$&1 \\
OnlyPublic &1 & 1 \\
OnlyPrivate &0 &$\min\{\frac{R}{\|\g_i\|},1\},\cdots$\\
\bottomrule[1pt]
\end{tabular}
\end{table}

\begin{theorem}[A closed-form solution for the optimal $\alpha$, informal]\label{theo:convex_alpha}
	Assume $n_{pub}$ public samples and $n_{priv}$ samples are drawn i.i.d. from the  distribution $\mathcal{D}$ and the loss objective for combining public and private gradients with weight $\alpha$ is convex in $\mathcal{W}$.
    The model parameters are denoted as $w$ with $w^*$ as the optimum point.
    Then, when achieving $\mu$-GDP~\cite{dong2022gaussian} at the limit with sufficiently large number of iterations $T$, the optimal $\alpha^*$ for balancing the optimization error and generalization error is
	$$
	\alpha^* = \mathop{\arg \min}\limits_{\alpha} \left[\frac{||\w_1-\w^*||(1-\alpha)\sqrt{d}}{\mu n_\textup{priv}} + \sqrt{\frac{(1-\alpha)^2}{n_\textup{priv}} + \frac{\alpha^2}{n_\textup{pub}} } \cdot \sqrt{\text{Var}_{z\sim \mathcal{D}}[l(z)]} \cdot \sqrt{d_\text{vc}} \right],
	$$
\end{theorem}
which is a function of $n_\textup{priv}, n_\textup{pub}$ the model dimension $d$, the VC dimension 
$d_\text{vc}$ for $\mathcal{W}$, and the privacy level $\mu$ (and thus the noise multiplier $\sigma$).

However, we show in \Cref{theo:convex_alpha} that choices of $\alpha$ in above methods are suboptimal for a convex setting.
Thus, we are motivated to formulate $\alpha$ for non-convex setting from a convergence perspective in \Cref{sec:convergence}, and therefore taking the batch size $B$, the model size $d$, the sample size $n$, number of iterations $T$, the privacy budget $\epsilon$, etc. into consideration.

\begin{remark}
With the post-processing property of DP, it is clear that \textit{Coupling} provides the same privacy guarantee as DP training with only private data when $\alpha > 0$.
When $\alpha = 0$, \textit{Coupling} degrades to conventional training on public data without privacy concerns. 
In other words, the choice of $\alpha$ does not affect the upper bound we obtained.
\end{remark}

\section{Convergence of new private training}\label{sec:convergence}
In this section, we analyze the convergence of \textit{Coupling} optimization for non-convex, positive and Lipschitz smooth loss, under standard assumptions used in the SGD literature.

\begin{assumption}\label{assumption-2smooth}
\label{assumption: Lipschitz}
(Smoothness). %
Let $\g(\w)$ denote the gradient of the loss $\mathcal{L}(\w)$, that is $L$-gradient Lipschitz such that $\forall \w, \mathbf{v}$,
$$
	\mathcal{L}(\mathbf{v}) - [\mathcal{L}(\w) + g(\w)^\top (\mathbf{v}-\w)] \le \frac{L}{2} ||\w-\mathbf{v}||^2.
$$
\end{assumption}

\begin{assumption}\label{assumption-3gradnoise}
(Gradient noise). 
The per-sample gradient noise $\g_{t,i}-\g_t$ is i.i.d. from a symmetric distribution such that
$$
\E(\g_{t,i}-\g_t) = 0, \E||\g_{t,i}-\g_t||^2 \le \xi^2.
$$
\end{assumption}

We emphasize that \Cref{assumption-3gradnoise} is commonly used in the non-DP SGD literature \cite{chen2020understanding,bu2022automatic,mandt2017stochastic,smith2018don,xie2020diffusion}. 

\subsection{Convergence Analysis}
Now we analyze the convergence
of the mixed training, using $\lesssim$ to denote the asymptotic inequality.
\begin{theorem}
\label{thm:norm converge}
    Under \Cref{assumption-2smooth} and \Cref{assumption-3gradnoise}, running DP-SGD for $T$ iterations and setting the learning rate $\eta \propto 1/\sqrt{T}$, with arbitrarily small probability $c>0$,
    $$
    \min_{0\leq t\leq T} \mathbb{P}\left(\|\g_t\|>f_r^{-1}\Big(\frac{2\cL_0+\frac{\xi^2}{2BL}}{c\sqrt{T}};\sigma,\cdots\Big)\right)\lesssim c
    $$
    where $f_r(\|\g\|;\sigma,\xi,d,B,\cL_0,L)$ is a positive function, increasing in $\|\g\|$ and decreasing in $\sigma$.
\end{theorem}

\Cref{thm:norm converge} indicates that, with high probability as $T\to\infty$, $f_r^{-1}$ and thus the gradient norm converges to zero asymptotically. 
In fact, the speed of convergence is negatively related to the initial loss $\cL_0$, the gradient variance $\xi$, and the noise multiplier $\sigma$ (which is inversely related to the privacy risk). 
We defer a more quantitative analysis on $f_r$ to the next section.

Note that hyperparameters such as $B, T, \sigma$ not only affect the convergence but also the privacy accounting. To disentangle the influence on the privacy and the accuracy, we study the convergence in \Cref{thm:norm converge} under a fixed privacy budget $(\epsilon, \delta)$. 

\begin{corollary}
\label{cor:mu}
Taking $\sigma^2$ equal to $1/\ln(\frac{n_\textup{priv}^2\mu^2}{TB^2} + 1)$, \Cref{thm:norm converge} stands for $\mu$-GDP. %

\end{corollary}

\subsection{Approximate analysis on the convergence}\label{subsec:approximateAnalysis}
To better understand the convergence, we simplify the inverse function $f^{-1}_r$ with \Cref{lam:function_fr}.
\begin{lemma}\label{lam:function_fr}
	For $T\to \infty$, the approximation for the inverse function in \Cref{thm:norm converge} is 
	$
	f_r^{-1} \approx \frac{\sqrt{4\cL_0 L+\frac{\xi^2}{B}}}{c^{1/2}T^{1/4}}(1-x)+o(x),
	$
	where $x=\sqrt{\frac{\cL_0 L n_\textup{priv}^2\mu^2}{2T d}}\frac{1}{4\xi^2}\to 0$.
\end{lemma}

Thus, we can expect the convergence to improve with the following guidelines, which are empirically validated by our experiments in \Cref{fig:effect}, \Cref{fig:heat} and \Cref{fig:com}:
\begin{itemize}
    \item Train longer with larger noise: fixing the batch size, model size and privacy budget, the bound $f_r^{-1}$ is smaller with a larger $T$ (though $\sigma$ is also larger).
    \item Larger batch size reduces the term $\frac{\sqrt{4\cL_0 L+\frac{\xi^2}{B}}}{c^{1/2}T^{1/4}}$. 
    \item Pre-training could lead to smaller $\xi$ that benefits the $\frac{\sqrt{4\cL_0 L+\frac{\xi^2}{B}}}{c^{1/2}T^{1/4}}$ term.
    \item Looser privacy budget $\mu$ and larger private data size $n_\textup{priv}$ increases $x$ which reduces the bound.
\end{itemize}

\subsection{Optimal $\alpha$ beyond public data-only training}
Previous works~\cite{golatkar2022mixed,amid2022public} have shown that introducing public data helps the convergence than private data-only training, which is reasonable because privacy is relaxed for public samples.
However, the gain over public-data-only training is still unknown.
We firstly compare \textit{Coupling} with \textit{OnlyPub}, which can be derived if we set $\eta_\textup{priv}=0$.
$$
\min_{0\leq t\leq T}\E(||\g_t||^2) \leq \E(\frac{1}{T}\sum_i^{B}\|\g_{t, i}\|^2)\leq \frac{1}{\sqrt{T}}[2\mathcal{L}_0 L+\frac{\xi^2}{B}]$$
Again by the Markov's inequality, for any $c\in(0, 1]$,
$$\min \mathbb{P}(\|\g\|^2> \frac{1}{c\sqrt{T}}[2\mathcal{L}_0 L+\frac{\xi^2}{B}])\leq c.$$

Given that $\xi$ is large, the comparison between \textit{Coupling} and non-DP SGD is roughly equivalent to that between
$\frac{\xi}{B^{1/2}c^{1/2}T^{1/4}}(1-x)$ and $\frac{\xi}{B^{1/2}c^{1/2}T^{1/4}}$. The advantage of private data is mainly in the sample size $n$ that increases $x$. This comparison shows that training with only public data (i.e. setting $\alpha=1$) is not optimal. 
As a consequence, we give Corollary \ref{cor:alpha} to derive the optimal $\alpha$ as the ratio between the optimal $\eta_\textup{pub}$ and $\eta_\textup{priv}$, and further validate it in \Cref{sec:optimal a} experiments.

\begin{corollary}
\label{cor:alpha}
Denote the learning rate for public and private data as $\eta_\textup{pub} = \eta \alpha$ and $\eta_\textup{priv} = \eta (1-\alpha)$, respectively. 
And the total dataset size is $n = n_\textup{pub} + n_\textup{priv}$, with the public data ratio as $r_\textup{pub} = \frac{n_\textup{pub}}{n_\textup{pub} + n_\textup{priv}}$.
Then under \Cref{thm:norm converge}, we suggest to apply the following $\alpha$ for better convergence
\begin{align}
    \alpha &= \frac{\eta_\textup{pub}}{\eta_\textup{priv} + \eta_\textup{pub}} = \frac{1}{1 + B \cdot \sqrt{\frac{2L\cL_0}{B^2 + \sigma^2d }}
    }.
\end{align}
We note that:
\begin{enumerate}
    \item $\alpha$ is monotonic decreasing with $B, L, \cL_0$ and $n_\textup{priv}$
    \item $\alpha$ is monotonic increasing with $\sigma, d, n_\textup{pub}$ and $r_\textup{pub}$
\end{enumerate}
\end{corollary}
Given that $\sigma\propto \frac{T}{\mu^2}$ by \eqref{eq:mu GDP}, we claim that $\alpha$ is also increasing with the number of iterations $T$ and decreasing with the privacy budget $\mu(\epsilon)$.
The evidence for the relation between $\alpha$ and $r_\textup{pub}$ (or equally $n_\textup{pub}$ and $n_\textup{priv}$) is shown in \Cref{fig:effect}.

It should be noted that we do not rely on calculating $\alpha$ with the exact $L$ and $\cL_0$ because they are fixed with the given task.
Instead, we can efficiently tune for an optimal $\alpha$ by applying \Cref{cor:alpha} with fewer grid search, thus improving the utility-privacy trade-off when providing additional privacy guarantee on hyper-parameters~\cite{papernot2021hyperparameter}.

\section{Experiments}
In this section, we validate our theoretical results and explore beyond them for CV and NLP tasks.

\textbf{Tasks and Setup.}
For image classification tasks, we use benchmark datasets~\cite{deng2012mnist, cifar10} of MNIST, CIFAR10, and CIFAR100. We train CNN (4 layers; from random initialization), ResNet18 ~\cite{he2016residual}, vision transformers (ViT)~\cite{VITdosovitskiy2020image} and DeiT~\cite{DEITtouvron2021training} from ImageNet pretrained weights.
For the text classification tasks, we train over SST-2 and QNLI datasets~\cite{glue} by fine-tuning the distilled Roberta~\cite{Sanh2019DistilBERTAD}.

To simulate our setting of a small portion of \textit{labeled and in-distribution} public data, we split the original training dataset into $r_\textup{pub}=\{0.01, 0.02, 0.05, 0.1, 0.2\}$ ratio of public data and take the rest as private data after a random shuffle.
To be consistent with our analysis, we apply SGD for the image task and  AdamW~\cite{loshchilov2017decoupled} for the NLP task with a mini-batch instead of a full-batch in AdaMix.
It should be noted that we apply a folklore~\cite{abadi2016deep, bu2022automatic, li2021large, golatkar2022mixed} choice of $\delta=1e^{-5}$ for CV tasks and $\delta=1/n^{1.1}$ for NLP tasks.

\textbf{Baselines.}
We compare with related works that train over a mix of public and private data for general optimizers.
For AdaMix~\cite{golatkar2022mixed}, we implement the adaptive clipping and set $\alpha=0.5$.
Following their adaptive projection, we tune the last linear layer and decompose its weight metrics.
For DPMD~\cite{amid2022public}, we use their approximate version that was actually carried out in their experiments and tune the hyper-parameter $i$ in $1-\alpha_t=\cos(\pi t/(Ti))$ with $i\in\{2, 3, 4, 5, 8\}$.
We also include non-mixed training methods of \textit{OnlyPriv},\textit{FullPriv}, \textit{OnlyPub} and \textit{NonPriv} as baselines.
\textit{FullPriv} and \textit{NonPriv} optimize the model over the whole training set with DP and non-DP optimizers respectively.

Previous works have indicated that a warm-up training with public data improves the performance~\cite{tramer2020differentially,abadi2016deep,bu2022automatic,li2021large,yu2021differentially}.
Therefore, we study the effect of warm-up~\cite{tramer2020differentially, golatkar2022mixed} in the last part of the experiments, though warm-up is not included in the main experiments for the fair comparison.
Besides, we also discuss the compatibility of \textit{Coupling} with other adaptive tricks in \textit{AdaMix}.

\subsection{Performance Evaluation}
We compare the accuracy of \textit{Coupling} with all baselines on various tasks in Table \ref{tab:acc_main}. \\
\textbf{A toss-up between only public or private training.}
For only private training, we can see that \textit{FullPriv} achieves higher accuracy than \textit{OnlyPriv} due to a larger training set.
On the one hand, \textit{OnlyPub} outperforms \textit{FullPriv} even if the amount of public data is much smaller (e.g., $r_\textup{pub}=0.05$), especially for an easier task with a smaller model capability (e.g., CNN model trained on MNIST).
A larger $r_{pub}$ enlarges the advantage of \textit{OnlyPub} over \textit{FullPriv}.
On the other hand, \textit{FullPriv} with more training samples can achieve a superior performance to \textit{OnlyPub} for harder tasks (e.g., CIFAR100).
The contrary observation validates that it is challenging to determine whether training on a larger private dataset with DP or training on a smaller public dataset without DP would result in a higher accuracy.
Generally speaking, our \textit{Coupling} performs better than both \textit{OnlyPub} and  \textit{FullPriv}, on par with the performance upper bound of \textit{NonPriv} across different tasks.

\begin{table*}[t]
\centering
\small
\caption{Accuracy comparison between FullPriv, NonPriv, OnlyPub, OnlyPriv and mixed training methods (AdaMix, DPMD and ours) under variant $r_\textup{pub}$ with $\epsilon=2$.}\label{tab:acc_main}
\resizebox{\textwidth}{!}{
\setlength\tabcolsep{1.2pt}
\begin{tabular}{c|c|c|ccccc|ccccc}
\toprule[1.5pt]
\multirow{2}{*}{Task} & \multirow{2}{*}{FullPriv} & \multirow{2}{*}{NonPriv} & \multicolumn{5}{c|}{$r_\textup{pub}$=0.05}                                     & \multicolumn{5}{c}{$r_\textup{pub}$=0.2}                                                                 \\
                      &                           &                          & OnlyPub & OnlyPriv & AdaMix & DPMD  & Ours                            & OnlyPub & OnlyPriv & AdaMix & DPMD                            & Ours                            \\ 
\midrule[1.5pt]
MNIST CNN             & 95.27                     & 98.96                    & 95.58   & 84.02    & 73.82  & 97.56 & \textbf{97.92} & 98.00   & 97.05    & 74.58  & 98.36                           & \textbf{98.36} \\
CIFAR10 ResNet18      & 58.76                     & 91.46                    & 52.80   & 52.09    & 44.33  & 62.45 & \textbf{65.09} & 70.36   & 49.86    & 46.55  & 70.71                           & \textbf{71.72} \\
CIFAR10 vit-small     & 97.11                     & 98.40                    & 95.87   & 96.96    & 89.10  & 97.12 & \textbf{97.39} & 97.29   & 96.81    & 90.60  & 97.66                           & \textbf{97.84} \\
CIFAR10 deit-small    & 93.90                      & 97.43                    & 92.92   & 93.79    & 90.02  & 94.02 & \textbf{95.55} & 95.93   & 93.39    & 90.91  & 95.12                           & \textbf{96.68} \\
CIFAR100 vit-small    & 82.98                     & 90.00                    & 67.66   & 77.33    & 77.17  & 82.03 & \textbf{84.41} & 86.62   & 73.42    & 81.21  & 86.54                           & \textbf{88.01} \\
CIFAR100 deit-small   & 65.88                     & 82.86                    & 30.99   & 45.23    & 63.03  & 69.69 & \textbf{72.42} & 72.48   & 39.28    & 68.97  & 77.90                           & \textbf{79.16} \\
SST2 distillRoberta   & 86.93                     & 92.66                    & 90.37   & 87.04    & 81.08  & 90.48 & \textbf{91.17} & 90.94   & 86.58    & 84.17  & \textbf{91.28} & 91.17                           \\
QNLI distillRoberta   & 82.78                     & 90.87                    & 90.37   & 81.88    & 64.76  & 84.28 & \textbf{91.06} & 84.40   & 82.74    & 65.62  & 87.59                           & \textbf{87.88} \\
\bottomrule[1.5pt]
\end{tabular}
}
\end{table*}

\textbf{Advantage of mixed training.}
From Table \ref{tab:acc_main}, we find that the mixed training, e.g. \textit{DPMD} and \textit{Coupling}, can achieve higher accuracy than non-mixed training, thus highlighting the importance of leveraging both public and private data.
Without the warm-up training, \textit{AdaMix} is inferior to non-mixed training on CIFAR10 and MNIST, possibly due to the information loss of the adaptive projection and only training the last layer.
Among all mixed training, \textit{Coupling} has the best performance, especially when the public data are scarcer ($r_\textup{pub}=0.05$) in comparison to $r_\textup{pub}=0.2$.
We emphasize that \textit{Coupling} does not need warm-up training, adaptive clipping, adaptive projection, nor the cosine scheduling, as long as the weight $\alpha$ is well-chosen by considering other parameters. 
This indicates that $\alpha$ is the key knob to improve the performance of the mixed training. Nevertheless,
we will show the compatibility of \textit{Coupling} to these tricks in a later section, where further improvement is observed.

\begin{figure}[htbp]
\centering
\includegraphics[width=0.4\textwidth, trim=0 0 0 0, clip]{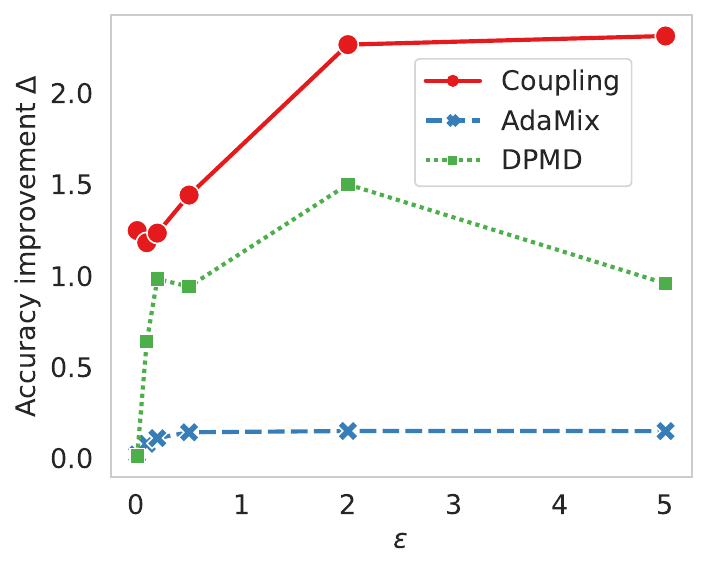}
\includegraphics[width=0.4\textwidth, trim=0 0 0 0, clip]{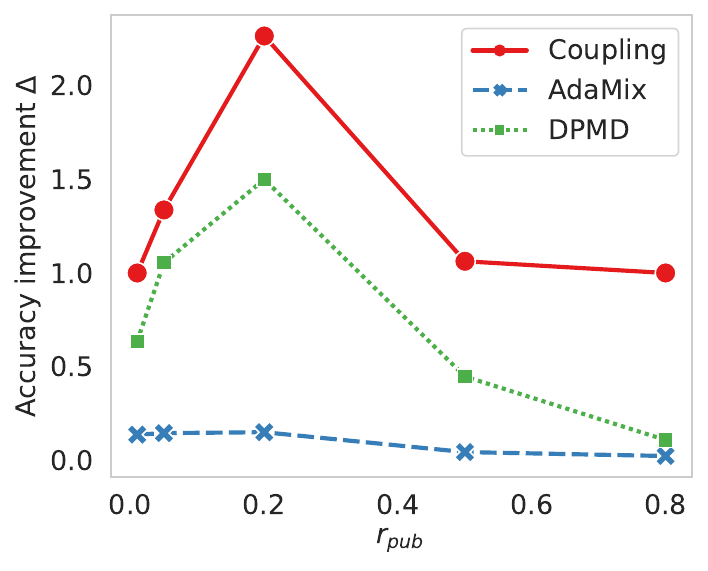}
\caption{Accuracy improvement $\Delta$ on ViT-small with CIFAR10 with $r_\textup{pub}=0.2, \epsilon=2$ by default.}
\label{fig:delta}
\end{figure}
To evaluate the mixed training, we follow the performance measure in \cite{golatkar2022mixed} as $\Delta = (\textup{acc}_{\textup{NonPriv}} - \max \{\textup{acc}_{\textup{OnlyPub}}, \textup{acc}_{\textup{FullPriv}}\}) / (\textup{acc}_{\textup{NonPriv}} - \textup{acc}_{\textup{method}})$, in which we consider the performance lower bound as the maximum accuracy of \textit{OnlyPub} or \textit{FullPriv}.
A larger $\Delta$ means a larger improvement boost by using the mixed training and the performance is approaching the upper bound of fully \textit{NonPriv} training (as if all data are public).
$\Delta<1$ indicates a utility drop compared to the maximum performance of \textit{OnlyPub} and \textit{FullPriv}.
As shown in Figure \ref{fig:delta}, \textit{Coupling} can guarantee $\Delta \ge 1$ for various $\epsilon$ and $r_\text{pub}$, while \textit{DPMD} and \textit{AdaMix} even cause a slight utility drop.
The reason may be that the cosine scheduling in \textit{DPMD} or the fixed weight in \textit{AdaMix} cannot balance the catastrophic utility drop brought by an extremely noisy private gradient.

Especially, we observe that $\Delta$ is monotonically increasing with a larger privacy budget.
Because when $\epsilon$ is small, applying private gradient with a large amount of noise ruins the model convergence of mixed training compared to \textit{OnlyPub}.
In addition, $\Delta$ is first increasing and then decreasing with an increasing $r_\textup{pub}$.
We notice that when $r_\textup{pub}$ is small, \textit{FullPriv} dominates the maximum performance in the numerator of $\Delta$ as $\max \{\textup{acc}_{\textup{OnlyPub}}, \textup{acc}_{\textup{FullPriv}}\}$, while \textit{OnlyPub} is superior when $r_\textup{pub}$ gets larger.
When \textit{FullPriv} dominates, the numerator of $\Delta$ increases while the denominator $\textup{acc}_{\textup{NonPriv}} - \textup{acc}_{\textup{method}}$ decreases.
When \textit{OnlyPub} dominates, the numerator decreases and the denominator approximates the value of the numerator.
Hence, the trend of $\Delta$ in Figure \ref{fig:delta} is reasonable.

To summarize,  \textit{Coupling} suggests that balancing the public and private gradients with an optimal $\alpha$ can be more effective than other tricks in \textit{AdaMix} and \textit{DPMD}.

\begin{figure}[htbp]
\centering

\centering
\includegraphics[width=0.4\textwidth, trim=0 0 0 0, clip]{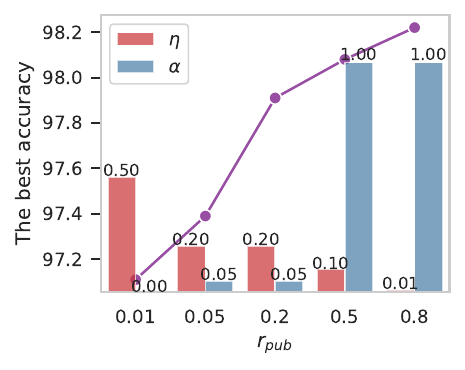}
\includegraphics[width=0.4\textwidth, trim=0 0 0 0, clip]{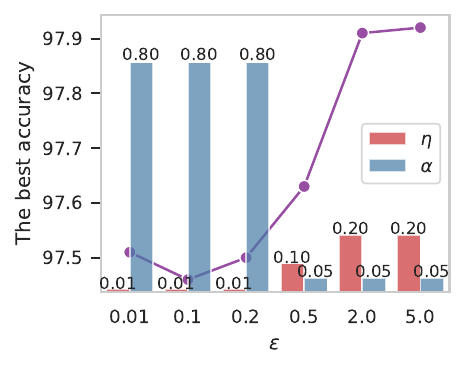}
\caption{Effects of $r_\textup{pub}, \epsilon$ on the best accuracy and the optimal choice of $\eta$ and $\alpha$ in \textit{Coupling} on ViT-small with CIFAR10.}
\label{fig:effect}
\end{figure}
\begin{figure*}[htbp]
\captionsetup[subfigure]{font=scriptsize,labelfont=scriptsize}
\centering
\subfigure[ViT-tiny]{
\label{subfig:vit-tiny}
\begin{minipage}[t]{0.27\linewidth}
\centering
\includegraphics[width=\textwidth, trim=65 45 65 110, clip]{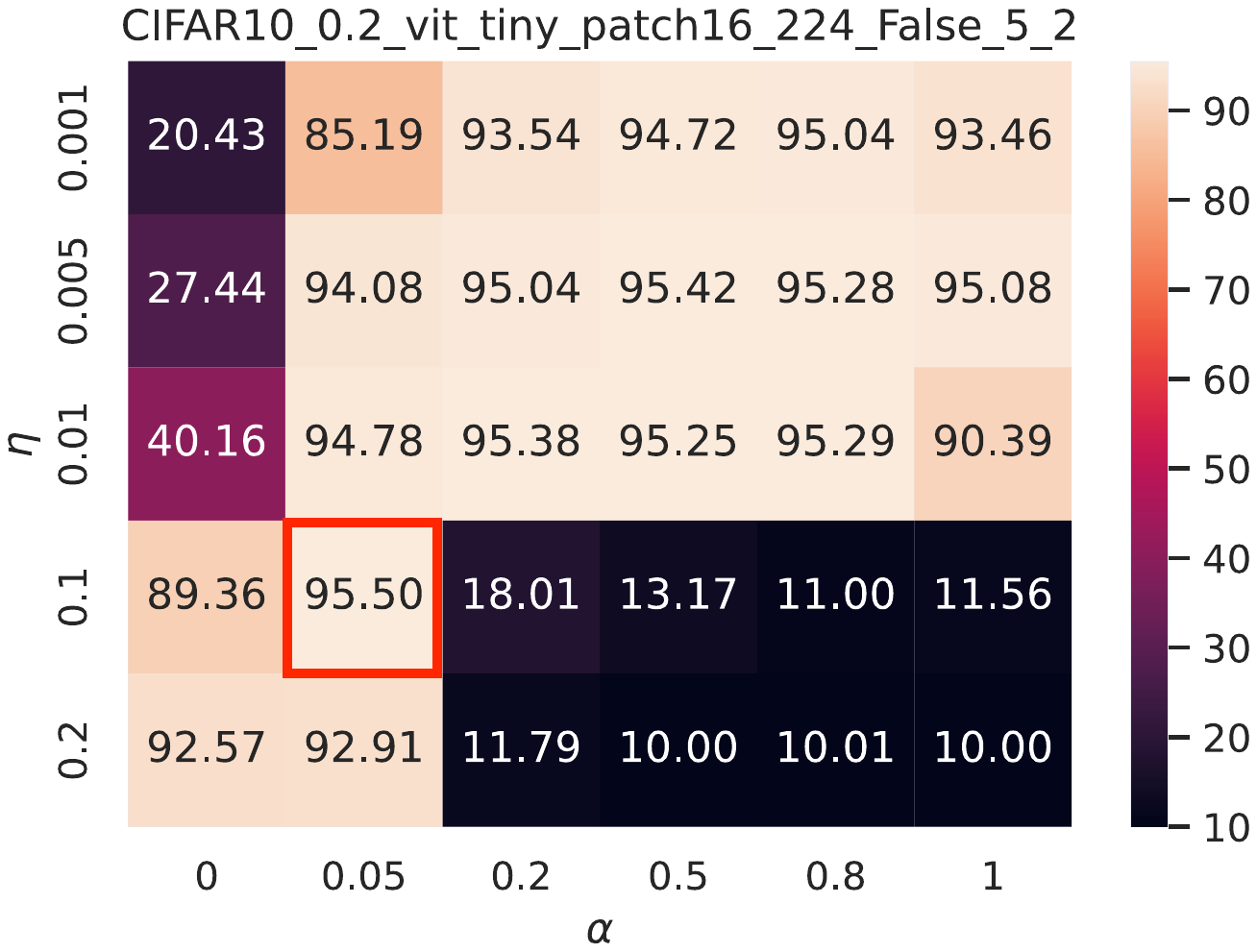}
\end{minipage}%
\hspace{-6mm}
}%
\subfigure[ViT-small]{
\label{subfig:vit-small}
\begin{minipage}[t]{0.27\linewidth}
\centering
\includegraphics[width=\textwidth, trim=65 45 65 110, clip]{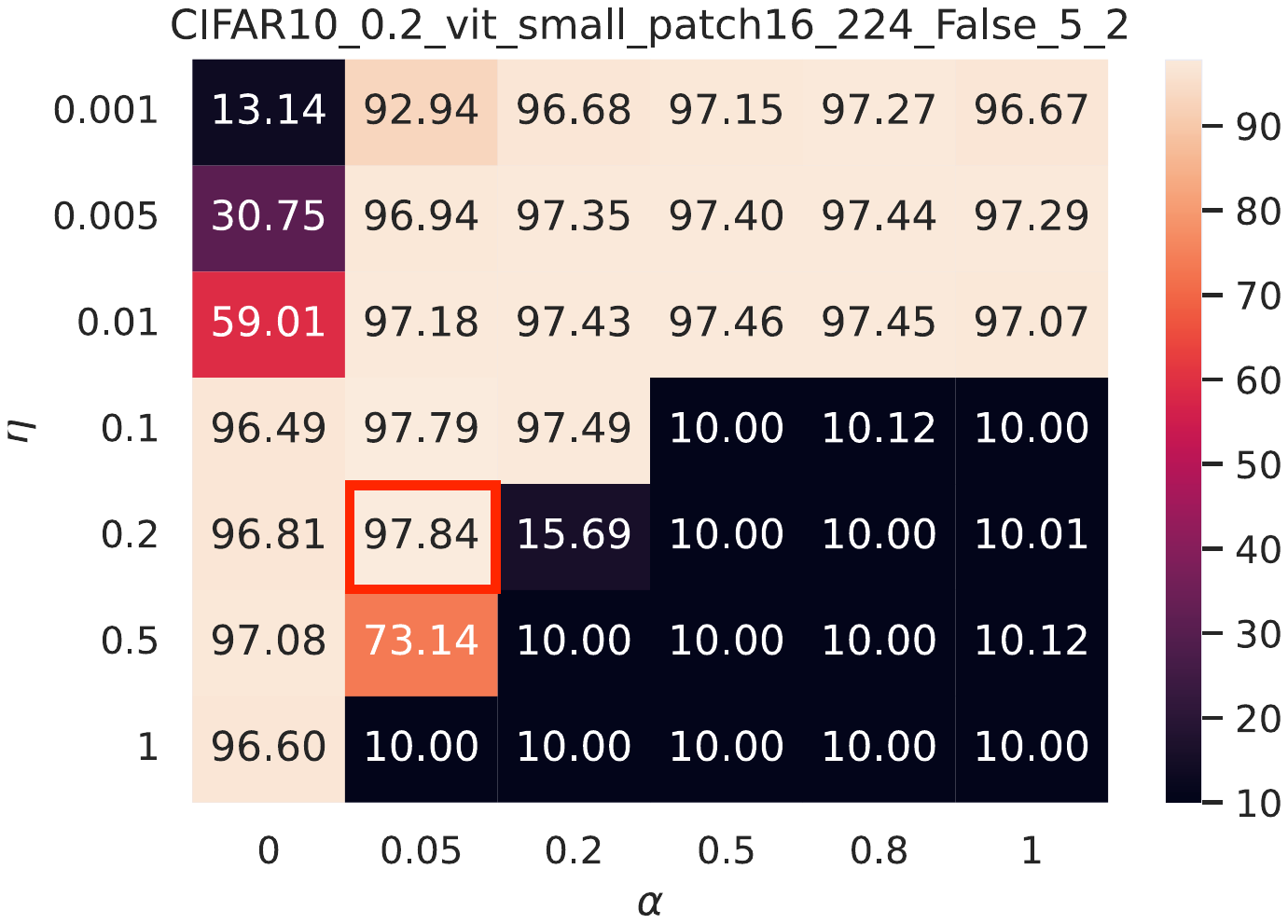}
\end{minipage}%
\hspace{-6mm}
}%
\subfigure[ViT-base]{
\label{subfig:vit-base}
\begin{minipage}[t]{0.27\linewidth}
\centering
\includegraphics[width=\textwidth, trim=65 45 65 110, clip]{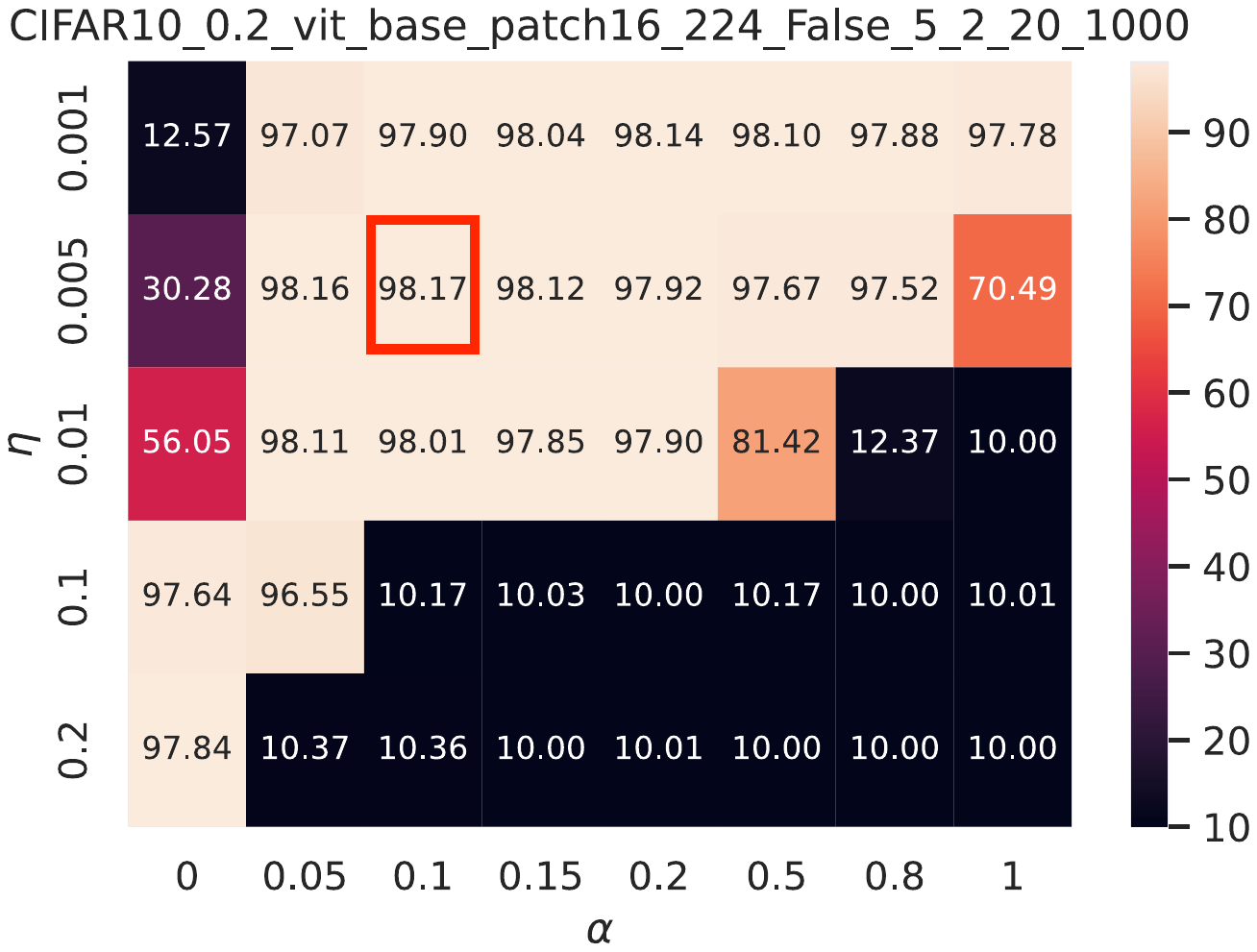}
\end{minipage}
\hspace{-6mm}
}%
\subfigure[ViT-large]{
\label{subfig:vit-large}
\begin{minipage}[t]{0.27\linewidth}
\centering
\includegraphics[width=\textwidth, trim=65 45 65 110, clip]{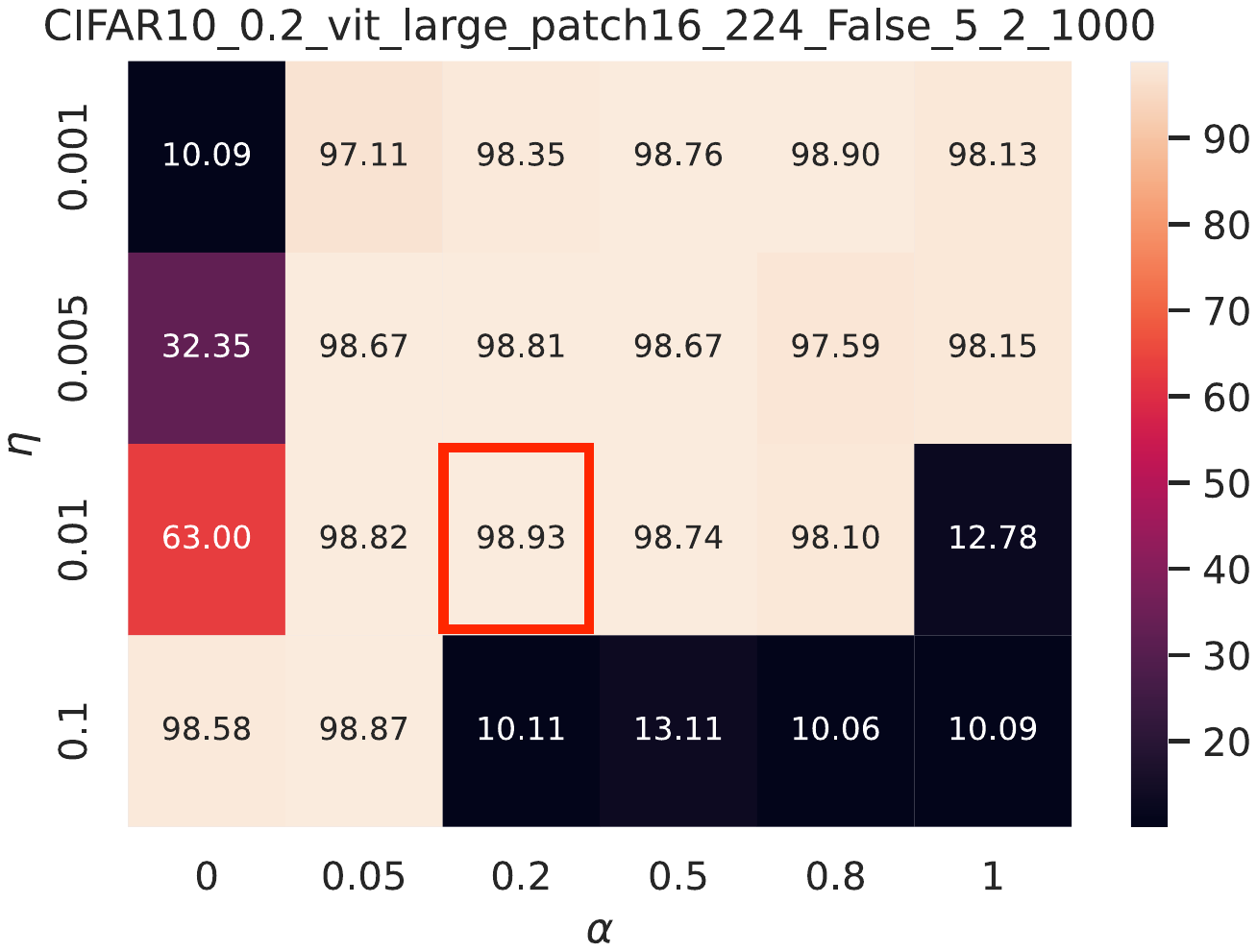}
\end{minipage}
\hspace{-6mm}
}%

\subfigure[ViT-small $B=100$]{
\label{subfig:b}
\begin{minipage}[t]{0.27\linewidth}
\centering
\includegraphics[width=\textwidth, trim=65 45 65 110, clip]{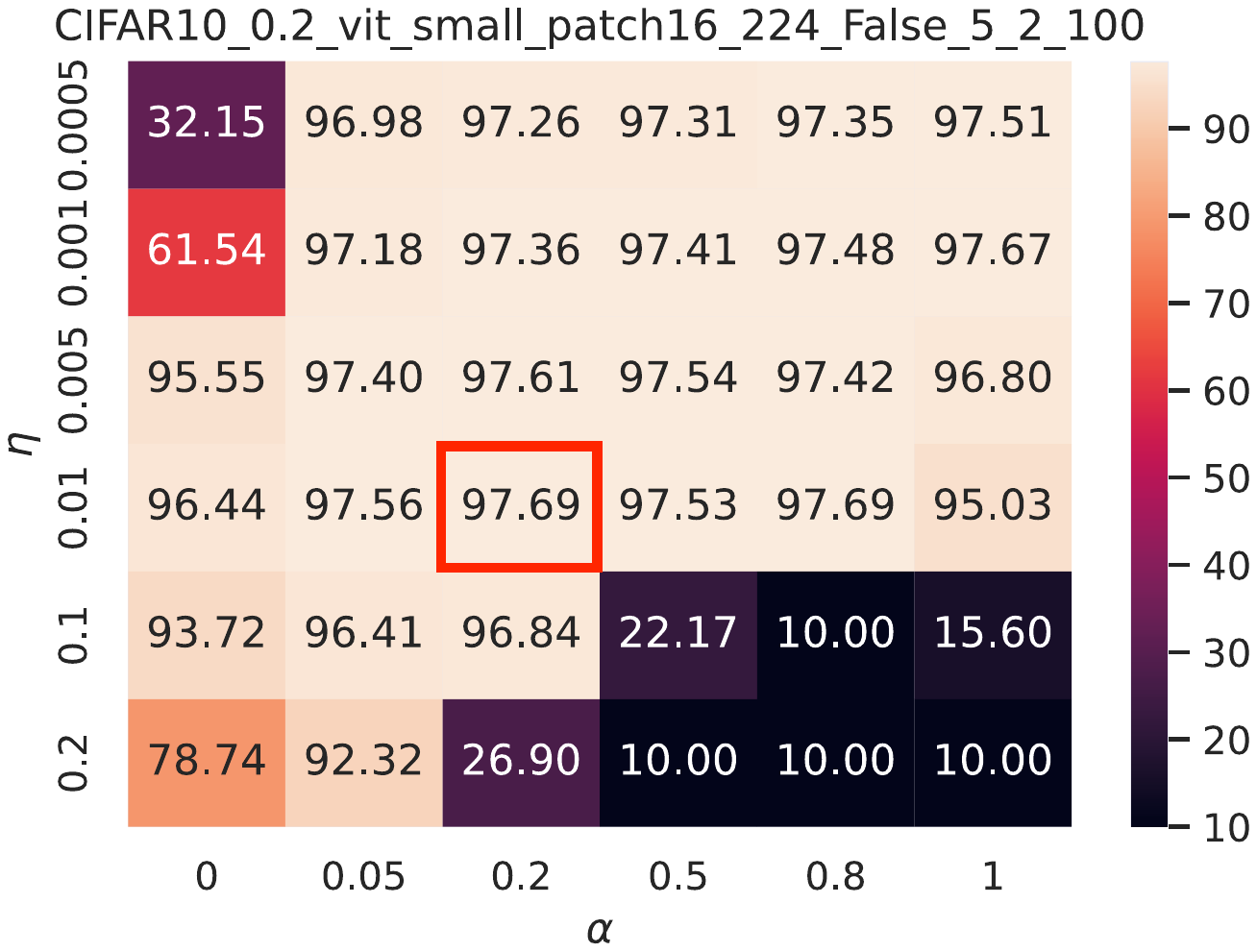}
\end{minipage}%
\hspace{-6mm}
}%
\subfigure[ViT-small $E=10$]{
\label{subfig:e}
\begin{minipage}[t]{0.27\linewidth}
\centering
\includegraphics[width=\textwidth, trim=65 45 65 110, clip]{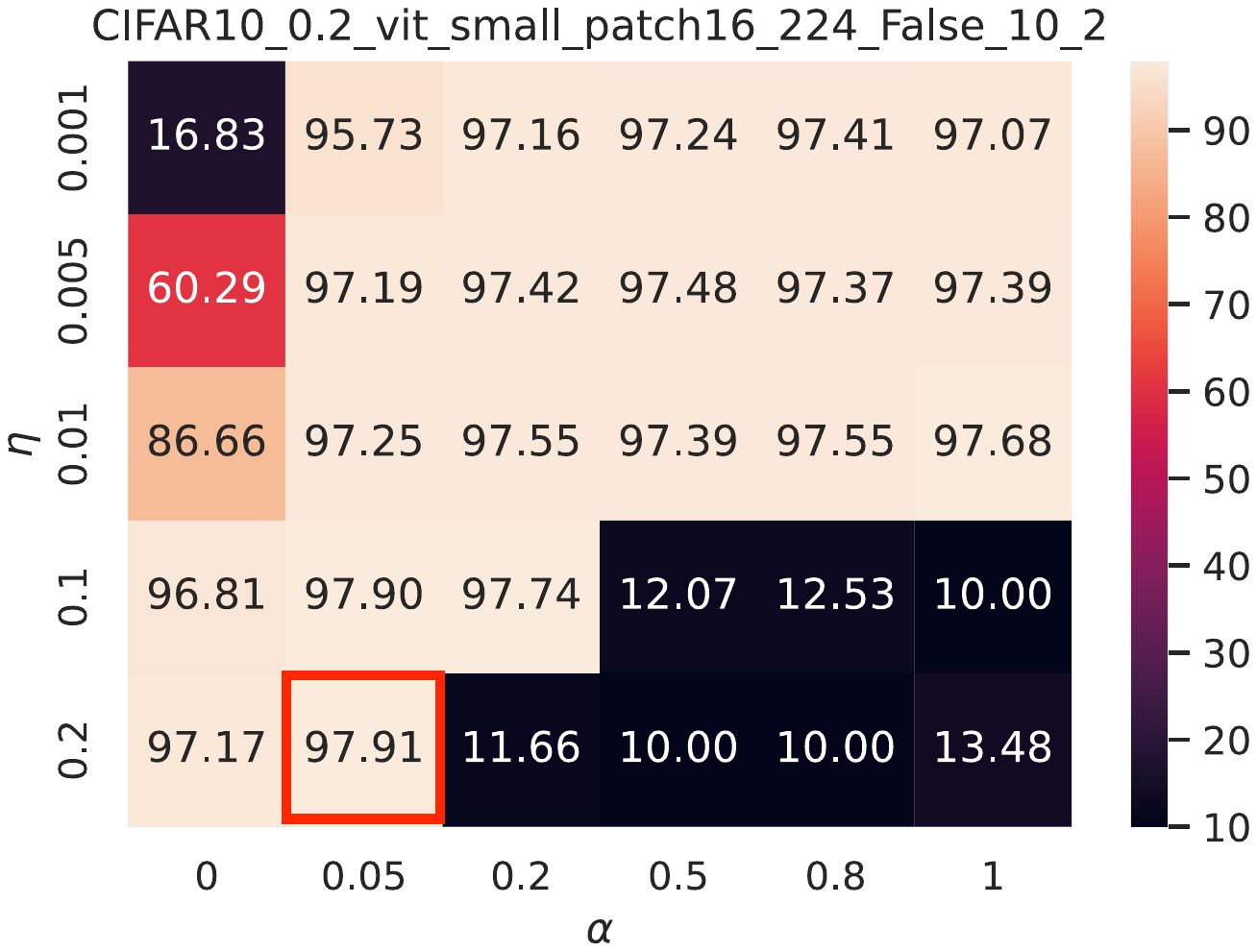}
\end{minipage}
\hspace{-6mm}
}%
\subfigure[ViT-small randomly initialized]{
\label{subfig:random}
\begin{minipage}[t]{0.27\linewidth}
\centering
\includegraphics[width=\textwidth, trim=65 45 65 110, clip]{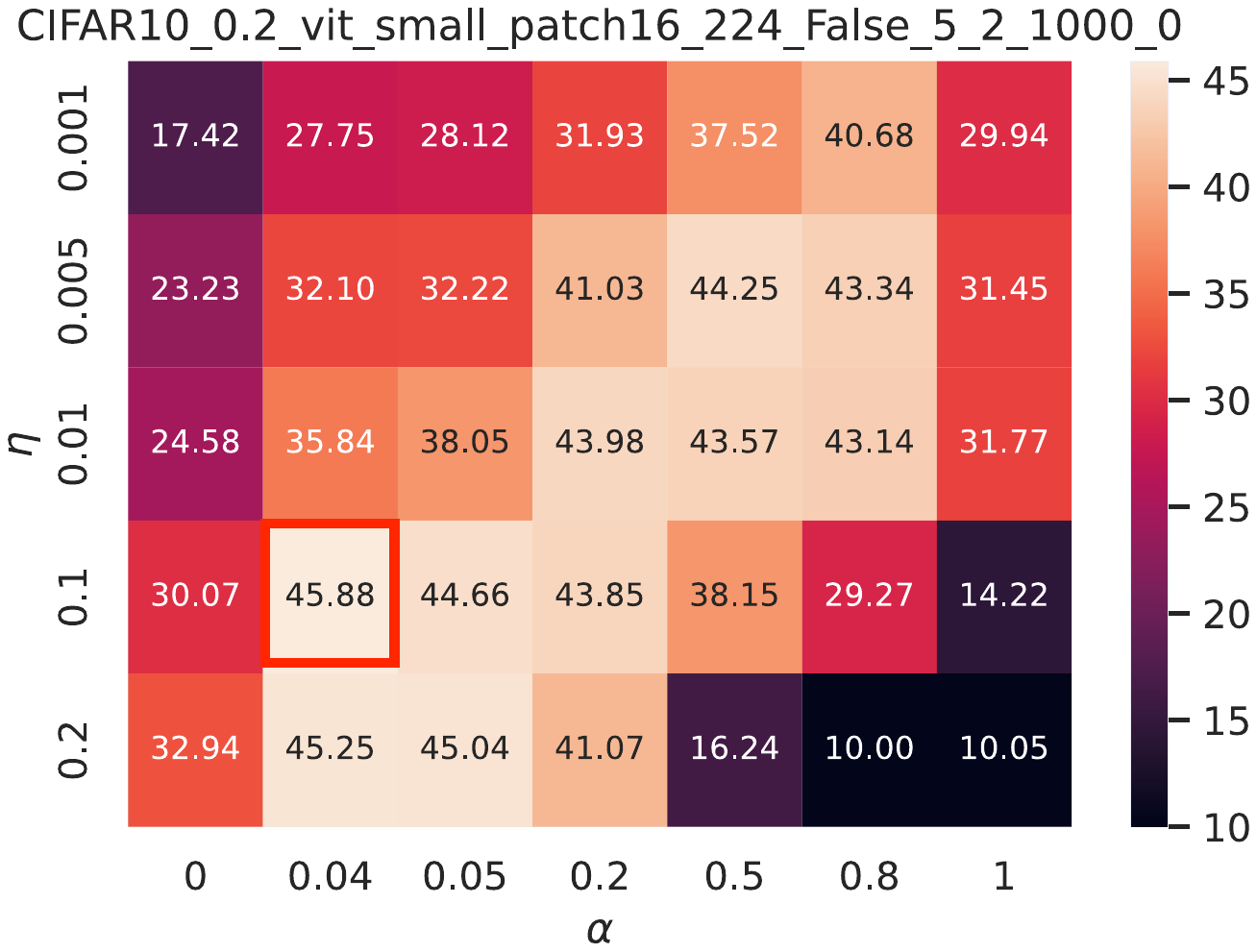}
\end{minipage}
\hspace{-6mm}
}%
\subfigure[DeiT-small]{
\label{subfig:deit}
\begin{minipage}[t]{0.27\linewidth}
\centering
\includegraphics[width=\textwidth, trim=45 45 45 100, clip]{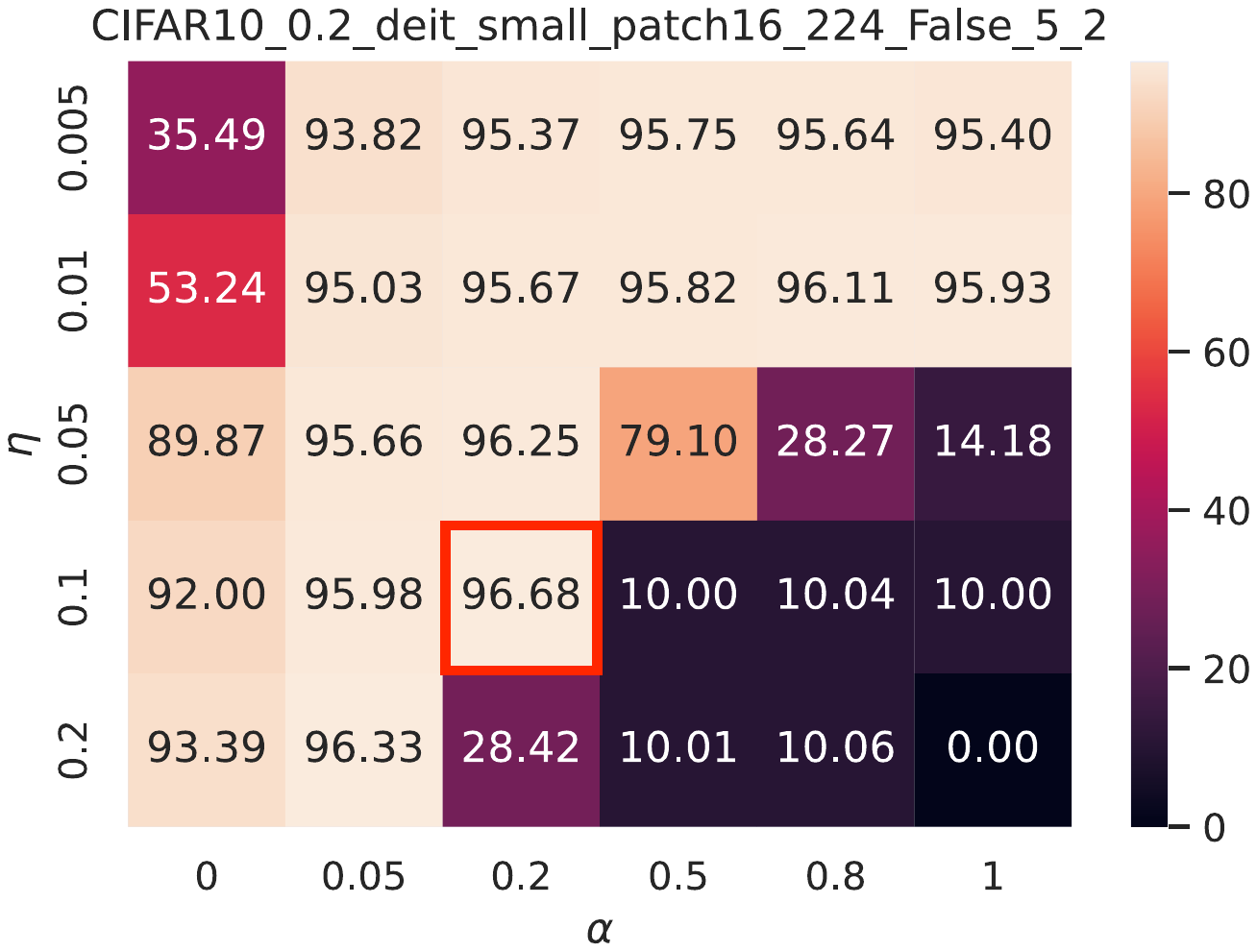}
\end{minipage}
\hspace{-6mm}
}%

\centering
\caption{Effect of $d, B, E, \epsilon$ and model structures on the optimal $\alpha$. 
The optimal accuracy is marked with a red rectangle.
By default, $\epsilon=2, r_\textup{pub}=0.2, E=5, B=1,000$.
}\label{fig:heat}
\end{figure*}

\subsection{Analysis of the Optimal $\alpha$}
\label{sec:optimal a}
To validate our analysis for the optimal $\alpha$ in \Cref{cor:alpha},
we present the heat map with X-axis of $\alpha$ and Y-axis of $\eta$ for \textit{Coupling} in \Cref{fig:heat} to show the effect of $r_\text{pub}$, $\epsilon$, $B$ and $d$ on the optimal choice of $\alpha$.
And we summarize the trend of optimal $\eta \in \{0.005, 0.01, 0.1, 0.2\}$ and $\alpha \in \{0.05, 0.2, 0.5, 0.8, 1\}$ for achieving the best accuracy in Figure \ref{fig:effect}.

\textbf{Effects of $r_\textup{pub}$ and $\epsilon$.}
From Figure \ref{fig:effect}, we observe that the optimal accuracy is increasing with a larger ratio of public data and a larger privacy budget.
For achieving the best accuracy with a larger $r_\textup{pub}$, $\alpha$ should be increased and $\eta$ should be decreased, which is reasonable because the accuracy gain of utilizing private data is reduced.
On the contrary, fixing $r_\textup{pub}=0.2$ and increasing the privacy budget, the best accuracy of \textit{Coupling} requires a larger learning rate $\eta$ and a smaller $\alpha$,
which echos Corollary \ref{cor:alpha} that the optimal $\alpha$ is increasing with a larger $\sigma$.

\textbf{Effects of $d, B$ and model structures.}
In Figure \ref{fig:heat}, we validate our claim in Corollary \ref{cor:alpha} that the optimal $\alpha$ is larger for a larger $d$ by comparing \Cref{subfig:vit-tiny}\ref{subfig:vit-small}\ref{subfig:vit-base}\ref{subfig:vit-large}, in which the optimal $\alpha$ ranges from $0.05$ to $0.2$.
As shown in \Cref{subfig:deit} and \Cref{subfig:vit-small}, different model structures also result in different optimal $\alpha$, which attributed to the term of $L$ and $\cL_0$ as shown in Corollary \ref{cor:alpha}.
We notice that in \Cref{subfig:random}, the optimal $\alpha$ for a ViT-small model\footnote{we omit the column of 0.04 in \ref{subfig:vit-small} as it does not increase the optimal accuracy.} which is initialized randomly (indicates a larger $\cL_0$) is 0.04, which reflects our conclusion in Corollary \ref{cor:alpha} that $\alpha$ is monotonic decreasing with $\cL_0$.
By comparing Figure \ref{subfig:b} and Figure \ref{subfig:vit-small}, it is obvious that we should increase $\alpha$ when the batch size $B$ is small to achieve the best accuracy.
In \Cref{subfig:e}, training for more epochs $E$ (equals to increasing the number of iterations $T$ and the noise magnitude $\sigma$) leads to better accuracy than \Cref{subfig:vit-small}, which validates our analysis in \Cref{subsec:approximateAnalysis}.
These observations provide a guideline to choose the optimal $\alpha$ for the best performance.

\subsection{Compatibility to training tricks}
\begin{figure*}[htbp]
\centering

\includegraphics[width=\textwidth, trim=0 0 0 0, clip]{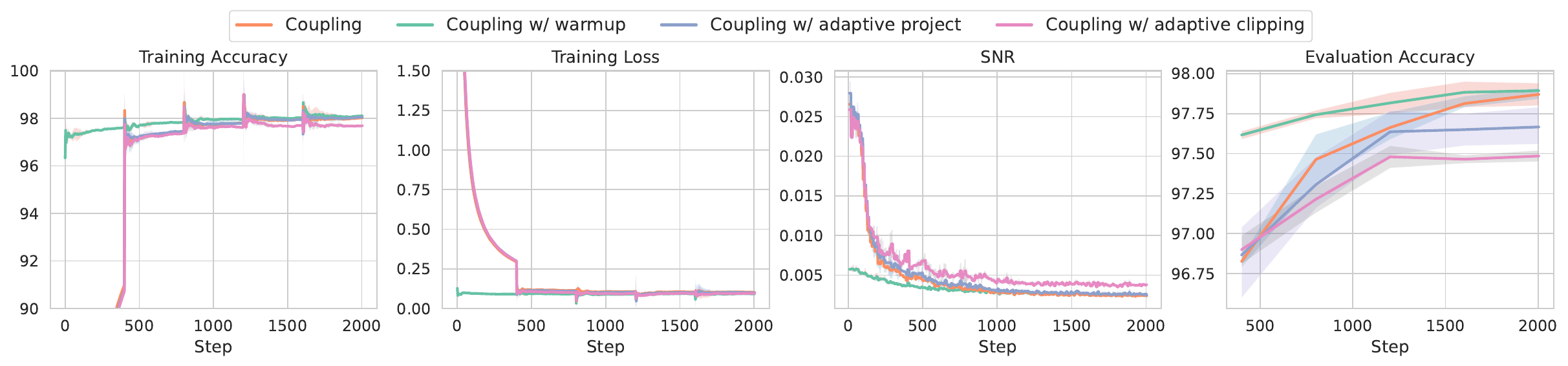}

\centering
\caption{Effects of warm-up, adaptive clipping, adaptive projection in \textit{Coupling} on CIFAR10 and ViT-small model with $\epsilon=2, r_{pub}=0.2$.}\label{fig:com}
\end{figure*}
We now equip \textit{Coupling} with the tricks in \textit{AdaMix} for further improvement.
As shown in \Cref{fig:com}, a warm-up training with a small amount of public data has faster convergence speed, higher accuracy, and smaller loss, though the performance is not significantly improved at the convergence.

As for the adaptive methods used in \textit{AdaMix},
we observe that \textit{Coupling w/ adaptive project} and \textit{Coupling w/ adaptive clipping} lead to a higher Signal-to-Noise (SNR), which is calculated as the ratio between the L2-norm of the clipped gradient sum for one batch and that of the injected noise vector.
On the contrary, the warm-up method leads to a lower SNR, which is caused by the fact that the gradient norm of a near-convergence model is small.
A higher SNR usually indicates a smaller magnitude of the perturbation. Yet, the evaluation accuracy of \textit{Coupling w/ adaptive clipping} and \textit{Coupling w/ adaptive project} is lower than \textit{Coupling} without these tricks, possibly due to the introduction of extra adaptive hyper-parameters, such as the compression ratio in adaptive clipping and the quantile in adaptive clipping.
In short, while \textit{Coupling} is compatible with additional tricks, the performance may be harmed or only improved marginally.

\section{Related works}
Previous works that consider both public and private data have two assumptions:
1) in-distribution public dataset, which follows the same distribution as the private dataset with a much smaller data size~\cite{bassily2018model, zhou2020bypassing, kairouz2020fast}; and 
2) out-of-distribution public datasets, which are usually easier to collect \cite{abadi2016deep, papernot2017semi, papernot2018scalable, li2021large} but more challenging to control the degree of domain shift.
In this work, we focus on the setting with a \textit{labeled and in-distribution} public dataset.

When using such public gradient implicitly in DP optimization, a line of works~\cite{kairouz2020fast, zhou2020bypassing, golatkar2022mixed, yu2021not}  make a low-rank assumption on the gradient subspace and utilize the structural information in public data to guide the training over private data.
\cite{kairouz2020fast} tracks historical gradients to do dimension reduction for private AdaGrad.
\cite{zhou2020bypassing} performs noise reduction by projecting the noisy gradients to a low-dimensional subspace, which is given by the top gradient eigenspace on a small public dataset.
However, gradients might not always hold the low-rank property, rendering these methods less accurate in practical settings.
Removing the low-rank assumption, AdaDPS\cite{li2022private} proposes to assist private training only for adaptive optimizers by accumulating gradients of the public dataset as the pre-conditioner, which avoids the accumulated noise in historical gradients.
In addition, the clipping threshold can be chosen adaptively~\cite{golatkar2022mixed, wang2020differentially, van2018three, zhang2022understanding, bagdasaryan2019differential, andrew2021differentially} based on the quantile of public gradient norms to mitigate the utility drop due to the clipping operation.

Recent works of mixed training~\cite{golatkar2022mixed,amid2022public} explicitly merge public gradients to help DP optimization.
Such linear combination of public and private components has also been studied for basic DP aggregation problems~\cite{ferrando2021combining,jorgensen2015conservative}.
We focusing on optimizing the linear combination for DP-SGD and our work can be compatible with existing mixed training methods and general optimizers.

\section{Discussion}
In this work, we investigate an effective way of mixed training by coupling public and private gradients with a linear combination, under the differential private regime.
Instead of choosing the weighting coefficient $\alpha$ as a constant in existing works, \textit{Coupling} formulates the optimal choice of $\alpha$ in a hyperparameter-dependent way.
As far as we know, this is the first work that provides a convergence analysis for mixed training in a non-convex setting, as well as a practical guideline to choose an $\alpha$ for better convergence based on training hyperparameters.
In addition, \textit{Coupling} allows better convergence than only public or only private training, approaching the utility upper bound of the model trained with full data in a non-DP way. 
We establish benchmark results to validate our theoretical analysis and show the compatibility of \textit{Coupling} with existing mixed training methods.

\bibliographystyle{unsrtnat}
\bibliography{arxiv_main}
\newpage
\appendix
\section{Experimental Details}\label{sec:appendixExp}
\subsection{Model Structure}
For CNN model on MNIST dataset in \Cref{tab:acc_main}, we follow the benchmark CNN model built in Tensorflow Privacy and Opacus, as shown below.
We set the same hyper-parameters as Tensorflow Privacy\footnote{https://github.com/tensorflow/privacy/tree/master/tutorials}.
The Non-Private baseline reaches $99\%$.
\begin{verbatim}
import torch.nn as nn
import torch.nn.functional as F

class CNN(nn.Module):
    def __init__(self, num_classes=10):
        super(CNN, self).__init__()
        self.conv1 = nn.Conv2d(1, 16, 8, 2, padding=3)
        self.conv2 = nn.Conv2d(16, 32, 4, 2)
        self.fc1 = nn.Linear(32 * 4 * 4, 32)
        self.fc2 = nn.Linear(32, num_classes)

    def forward(self, x):
        # x of shape [B, 1, 28, 28]
        x = F.relu(self.conv1(x)) # -> [B, 16, 14, 14] 
        x = F.max_pool2d(x, 2, 1) # -> [B, 16, 13, 13] 
        x = F.relu(self.conv2(x)) # -> [B, 32, 5, 5]
        x = F.max_pool2d(x, 2, 1) # -> [B, 32, 4, 4] 
        x=x.view(-1,32*4*4) #->[B,512]
        x = F.relu(self.fc1(x)) # -> [B, 32]
        x = self.fc2(x) # -> [B, 10]
        return x
\end{verbatim}

The model of distilled Roberta\cite{Sanh2019DistilBERTAD} can be found in this repository
\footnote{https://github.com/huggingface/transformers/tree/main/examples/research\_projects/distillation}.
For ResNet18 ~\cite{he2016residual}, vision transformers (ViT)~\cite{VITdosovitskiy2020image} and DeiT~\cite{DEITtouvron2021training} models, we apply existing pretrained models via timm\footnote{https://github.com/rwightman/pytorch-image-models}.

\subsection{Default Experimental Setup}
\textbf{Hyper-parameters for training.} We use the SGD optimizer for CV tasks and AdamW~\cite{loshchilov2017decoupled} for NLP tasks.
By default, we set $B=1,000$ for all tasks and the following parameters are listed in Table \ref{tab:append_hyperParam} for different tasks:
\begin{table*}[h!]
\centering
\caption{Hyper-parameters for different tasks.}\label{tab:append_hyperParam}
\resizebox{0.65\textwidth}{!}{
\begin{tabular}{c|cccc}
\toprule[1.5pt]
Task                & $\eta$ range                  & mini batch size & epochs \\
\midrule[1.5pt]
MNIST CNN           & 1e-2 5e-2 1e-1 1.5e-1 2e-1    & 200             & 20     \\
CIFAR10 ResNet18    & 5e-3 1e-2 5e-2 1e-1 5e-1      & 200             & 30     \\
CIFAR10 vit-tiny    & 1e-3 5e-3 1e-2 1e-1 2e-1      & 200             & 5      \\
CIFAR10 vit-small   & 1e-3 5e-3 1e-2 1e-1 2e-1      & 200             & 5      \\
CIFAR10 vit-large   & 1e-3 5e-3 1e-2 1e-1 2e-1      & 200             & 5      \\
CIFAR10 vit-base    & 5e-4 5e-3 1e-3 1e-2 1e-1 2e-1 & 200             & 5      \\
CIFAR10 deit-small  & 5e-3 1e-2 5e-2 1e-1 2e-1      & 200             & 5      \\
CIFAR100 vit-small  & 5r-4 1e-3 5e-3 1e-2 1e-1 5e-1 & 200             & 5      \\
CIFAR100 deit-small & 5r-4 1e-3 5e-3 1e-2 1e-1 5e-1 & 200             & 5      \\
SST2 distillRoberta & 1e-5, 5e-5, 2e-4, 5e-4        & 10              & 3      \\
Qnli distillRoberta & 5e-5, 2e-4, 5e-4              & 10              & 6  \\
\bottomrule[1.5pt]
\end{tabular}
}
\end{table*}

\textbf{Environment.}
Our experiments are conducted on a Linux
server with 8 Tesla V100-SXM2 GPUs with 16GB memory.

\section{Notation Statement}\label{sec:appendix:notation}
For better readability, we explain notations in \Cref{tab:notation}.
\begin{table*}[t]
\caption{Notation Statement}\label{tab:notation}
\centering
\begin{tabular}{l|l}
\toprule[1pt]
\textbf{Notation} & \textbf{Meaning} \\
\midrule[1pt]
$n$ & total number of the whole dataset with $n = n_{pub} + n_{priv}$ \\
$r_{pub}$ & ratio of public data in the whole dataset with $r_{pub}=n_{pub} / n$ \\
$B$ & batch size; we omit the subscript when $B_{pub}=B_{priv}$  \\
$t$ & $t^{\text{th}}$ iteration \\
$T$ & total number of iterations \\
$\g_{t, i}$ & per-sample gradient of the $i^{\text{th}}$ sample in the $t^{\text{th}}$ iteration \\
$C(\cdot)$ & clipping function \\
$C_{t, i}$ & exact clipping threshold for the $i^{\text{th}}$ sample in the $t^{\text{th}}$ iteration; \\
$R$ & clipping threshold; not necessary for automatic clipping \\
$\sigma$ & standard deviation of the noise distribution \\
$\g_t$ & gradient of a batch \\ %
$L$ & Lipschitz constant \\
$\cL$ & loss value over a batch of samples; $\cL_0$ as the initial loss \\
$\eta$ & learning rate; $\eta_{pub}$ for public gradient and $\eta_{priv}$ for private gradient \\
$\alpha$ & weight for public gradient in the linear combination of gradients \\
$d$ & number of model parameters \\
\midrule[1pt]
$(\epsilon, \delta)$ & privacy budget of the approximate differential privacy\\
$\mu$ & privacy parameter of the Gaussian differential privacy~\cite{bu2020deep} \\
\midrule[1pt]
$\xi$ & parameter of the gradient noise \\
$c$ or $a$ & a probability value \\
$r$ & any positive constant with $r>0$ \\
$\cL_r$ & a function of $\|\g\|$ conditioned by $c, \sigma,\xi,d,B,\cL_0$ and $L$; $f^{-1}$ is its inverse function \\
\midrule[1pt]
$\alpha_t$ & weight for public gradient in \cite{amid2022public} in the $t^{\text{th}}$ iteration \\ 
$i$ & hyper-parameter introduced in \cite{amid2022public} to tune $\alpha_t$, when not shown in any subscript \\
\bottomrule[1pt]
\end{tabular}
\end{table*}

\section{Discussion on the Choice of $(\epsilon, \delta)$}\label{sec:discussion_on_choice_eps}
It should be noted that the dataset size should be considered for choosing a proper $\delta$ in practice. 
Even though $\delta=1e^{-5}$ is widely adopted as a simplified choice for building benchmarks. 
In practice, we can follow a more principled way in choosing $\delta$. 
The key point here is that the choice of $\delta$ should inversely scales with the number of private samples in training. For example, a folklore choice\footnote{\url{https://github.com/lxuechen/private-transformers/blob/main/private_transformers/privacy_engine.py}} $\sigma = \frac{1}{n^{1.1}}$ for pure private training.

\begin{figure}[htbp]
\centering
\includegraphics[width=0.4\textwidth, trim=0 0 0 0, clip]{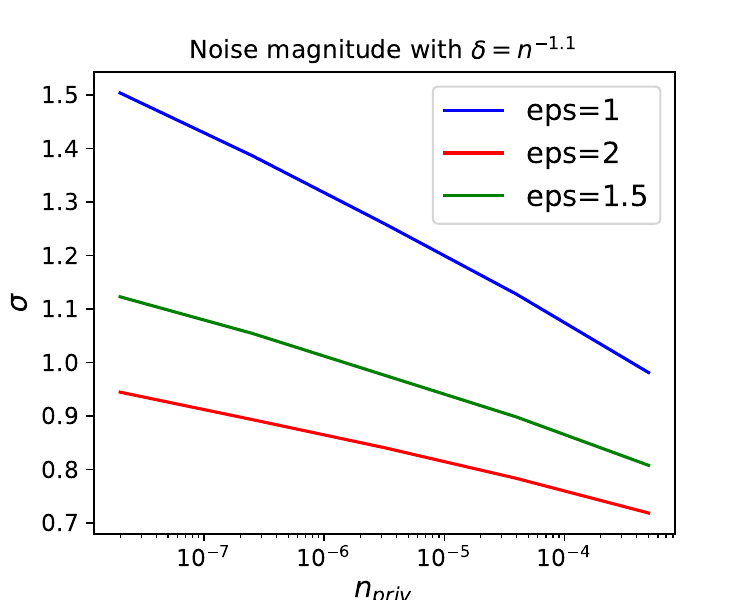}
\caption{Relation of $\sigma$ and $n_\text{priv}$ when setting $\delta=n^{-1.1}$}
\label{fig:eps_delta_n}
\end{figure}

\begin{table}[htb]
\centering
\caption{Privacy budget $\epsilon$ under a different $\delta$}\label{tab:epsilon}
\begin{tabular}{c|ccc}
\toprule[1pt]
Dataset       & $n_{\text{priv}}$     & $(\epsilon, \delta=n_\textup{priv}^{-1.1})$ & $\epsilon$ with $\delta=1e^{-5}$ \\
\midrule[1pt]
CIFAR10~(50,000)       & 47,500 ($r_{pub}=0.05$) & $(2.00001, 7.17e^{-6})$                   & 2                                       \\
              & 40,000~($r_{pub}=0.2$)  & $(2.00001,8.66e^{-6})$                    & 2                                       \\
              & 40,000~($r_{pub}=0.2$)  & $(5.00003,8.66e^{-6})$                    & 5                                       \\
              \hline
SST2~(67,349)  & 53,879~($r_{pub}=0.2$)  & $(2.00001,6.24e^{-6})$                   & 2                                       \\
\hline
QNLI~(104,743) & 83,794~($r_{pub}=0.2$)  & $(2.00001,3.84e^{-6})$                   & 2              \\
\bottomrule[1pt]                        
\end{tabular}
\end{table}

Essentially, the value of $(\epsilon, \delta)$ reflects the magnitude of noise $\sigma$ that we inject in each experiment. 
Given two variable in $\epsilon, \delta$ and $\sigma$, we can derive the rest one from a DP accountant oracle. 
Thus, for our current results where a known $\sigma$ is applied, if we would like to set $\delta$ after considering the private dataset size, we can map the privacy level $(\epsilon, 1e^{-5})$ in the current version with the newly derived $(\epsilon, n_\textup{priv}^{-1.1})$ for different number of private samples. 
In \Cref{tab:epsilon}, we demonstrate the privacy level for several combinations after considering the influence of private sample size. 
In \Cref{fig:eps_delta_n}, we show the scaling trend of noise magnitude with different size of public dataset $n_\text{priv}$ for different choices of $\epsilon$.
The same procedure can be performed for choosing $n_{\text{priv}}$-dependent privacy parameters $\epsilon$ and $\delta$.

\section{Main proofs}\label{sec:norm converge}
\subsection{Proof of \Cref{theo:convex_alpha}}\label{sec:proof of theo:convex_alpha}
\textit{Proof.}
	Denote the model parameters as $\w$ with $\w^*$ as the point where the optimum is attained.
The loss objective for combining public and private gradients with $\w$ is
\begin{align}
\cL_{\alpha}(\w) = \frac{\alpha}{n_{pub}}\sum_{i}^{n_{pub}}l_i +  \frac{1-\alpha}{n_{priv}}\sum_j^{n_{priv}}l_j.
\end{align}

The variance of the combined loss $\cL_\alpha(\w)$ gives us how much difference the objective function we have is approximating the stochastic objective function of interests:
\begin{align}
\text{Var}[\cL_{\alpha}(\w)] = \frac{\text{Var}[l_{z \sim \mathcal{D}_\text{pub}}]}{n_{pub}}\cdot\alpha^2 + \frac{\text{Var}[l_{z \sim \mathcal{D}_\text{priv}}]}{n_{priv}}\cdot  (1-\alpha)^2 \\
= \text{Var}[l_{z\sim \mathcal{D}}] \cdot \left( \frac{\alpha^2}{n_{pub}} + \frac{(1-\alpha)^2}{n_{priv}} \right).
\end{align}

In terms of generalization, we can derive an optimal $\alpha$ by minimizing the following term which describes how well $\cL_\alpha(\w)$ approximates the expected loss over the data distribution $\mathbb{E}_{z\sim \mathcal{D}}[l(z)]$ as:
\begin{align}
\alpha_\text{gen}^* 
&= \mathop{\arg \min}\limits_{\alpha}
\left(
\sqrt{\frac{(1-\alpha)^2}{n_{priv}} + \frac{\alpha^2}{n_{pub}} } \cdot \sqrt{\text{Var}_{z\sim \mathcal{D}}[l(z)]} \cdot \sqrt{\text{VCdim}} \right) 
= \frac{n_\text{pub}}{n_\text{pub} + n_\text{priv}}
\end{align}

In terms of optimization, an optimal $\alpha$ should result in a smaller variance of the combined loss $\cL_\alpha$ for a smaller optimization error i.e., $\cL_{\alpha}(\hat{\w_T}) - \cL_{\alpha}(\w^*)$.
We suppose $n_\text{pub}, n_\text{priv} \gg B$, thus the  variance of the stochastic gradient oracle in mixed training is approaching that of sampling with replacement.
For brief, we denote the variance related terms with $W^2$.
\begin{align}
\text{Var}[\triangledown \cL_\alpha(\w)] 
&= 
\text{Var}[\frac{1-\alpha}{B} \cdot \left( \sum_{i=1, i \sim D_\text{priv}}^B \triangledown l_i + \mathcal{N}(0, \sigma^2\mathbb{I}_d )\right) + \frac{\alpha}{B} \cdot \sum_{j=1, j\sim D_\text{pub}}^B\triangledown l_j] \nonumber \\
&= \frac{(1-\alpha)^2}{B} \cdot \text{Var}_{i\sim D_\text{priv}} \triangledown l_i + \frac{\alpha^2}{B} \cdot \text{Var}_{j\sim D_\text{pub}} \triangledown l_j + \frac{(1-\alpha)^2}{B^2} d\sigma^2 \nonumber \\
&\approx \frac{W^2}{B} + \frac{(1-\alpha)^2 d\sigma^2}{B^2} 
\end{align}
 
For convex $\cL_{\alpha}$ with $\eta \leq \frac{1}{L}$, we have the expected optimization error as follows
\begin{align}
E[\cL_{\alpha}(\hat{\w}_T)] - \cL_{\alpha}(\w^*) &\leq \frac{||\w_1 - \w^*||^2 + (\frac{W^2}{B} + \frac{(1-\alpha)^2 d \sigma^2}{B^2} ) \cdot T \cdot \eta^2}{T \eta}
\end{align}
Fixing the parameters $T, \sigma^2$ and $B$ to obtain $\mu$-GDP~\cite{dong2022gaussian}, we have $\frac{TB^2}{\sigma^2 n_{priv}^2} \asymp \mu^2 $, thus
\begin{align}
E[\cL_{\alpha}(\hat{\w}_T)] - \cL_{\alpha}(\w^*) &\leq \frac{||\w_1-\w^*||^2}{T \eta} + \frac{W^2\eta}{B} + \frac{(1-\alpha)^2d\cdot T\cdot \eta}{\mu^2\cdot n_{priv}^2}
\end{align}
The ideal choice of $\eta \to 0$ as $T\to \infty$, so the second term vanishes.
If we choose
$
T\eta  = 
\sqrt{\frac{||\w_1-\w^*||^2 \mu^2 n_{priv}^2}{(1-\alpha)^2d}}
$, the resulting $\frac{||\w_1-\w^*|| (1-\alpha) \sqrt{d}}{n_{priv} \mu}$ achieves the information-theoretic limit.
Thus, the optimal choice for optimization is $\alpha^*_\text{opt}=1$.

Combining the two views and trading off between minimizing the optimization error and the generalization error, we solve the optimal $\alpha^*$ as:
$$
\alpha^* = \mathop{\arg \min}\limits_{\alpha} \left[\frac{||\w_1-\w^*|| (1-\alpha) \sqrt{d}}{n_{priv} \mu} + \sqrt{\frac{(1-\alpha)^2}{n_{priv}} + \frac{\alpha^2}{n_{pub}} } \cdot \sqrt{\text{Var}_{z\sim \mathcal{D}}[l(z)]} \cdot \sqrt{\text{VCdim}} \right]
$$
In fact, the objective on the right hand side is convex in $\alpha$, because its second derivative is
$$\frac{\sqrt{n_{pub}n_{priv}}}{\left(n_{priv}\alpha^2+n_{pub}(1-\alpha)^2\right)^{3/2}}\cdot \sqrt{\text{Var}_{z\sim \mathcal{D}}[l(z)]} \cdot \sqrt{\text{VCdim}}>0.$$
Therefore, $\alpha^*$ is unique and satisfies
$$\frac{-\frac{(1-\alpha)}{n_{priv}} + \frac{\alpha}{n_{pub}}}{\sqrt{\frac{(1-\alpha)^2}{n_{priv}} + \frac{\alpha^2}{n_{pub}} } }\cdot \sqrt{\text{Var}_{z\sim \mathcal{D}}[l(z)]} \cdot \sqrt{\text{VCdim}}=\frac{||\w_1-\w^*|| \sqrt{d}}{n_{priv} \mu}$$
which can be derived from the unique stationary point of the objective. Solving this quadratic polynomial is ignored here. We note that the optimal $\alpha^*$ is a function of $n_\textup{priv}, n_\textup{pub}, d, \text{VCdim}$ and the privacy level $\mu$ (and thus the noise multiplier $\sigma$).

\subsection{Proof of \Cref{thm:norm converge}}\label{sec:proof thm:norm converge}
\textit{Proof.}
By Lipschitz smoothness in \Cref{assumption: Lipschitz} and denoting $\Z=\mathcal{N}(0,\I)$, we have
\begin{align*}
&\mathcal{L}_{t+1}-\mathcal{L}_t
\leq \g_t^\top (\w_{t+1}-\w_t)+\frac{L}{2}\|\w_{t+1}-\w_t\|^2
\\ 
&=-\g_t^\top\left(\eta_\text{pub}\sum_j^{B} \g_{t, j} + \eta_\text{priv} \left[\sum_i^{B} \frac{\g_{t, i}}{||\g_{t, i}|| + \gamma }+\sigma \Z\right]\right)\\
&+ \frac{L}{2}\left\| \eta_\text{pub} \sum_j^{B} \g_{t, j} + 
\eta_\text{priv} \left[ \sum_i^{B} \frac{\g_{t, i}}{||\g_{t, i}|| + 1 } + \sigma \Z \right] \right\|^2
\end{align*}
Taking the expectation over the randomness of sampling,
\begin{align*}
&\E(\cL_{t+1} - \cL_{t}|\w_t) \leq -B\g_t^\top\left(\eta_\text{pub} \g_{t} + \eta_\text{priv} \E\frac{\g_{t, i}}{||\g_{t, i}|| + 1 }\right)\\
&+ \frac{L}{2}\E\left\| \eta_\text{pub} \sum_j^{B} \g_{t, j} + 
\eta_\text{priv} \left[ \sum_i^{B} \frac{\g_{t, i}}{||\g_{t, i}|| + 1 } + \sigma \Z \right] \right\|^2
\end{align*}
We directly expand the $\E\|\cdot\|^2$ term above:
\begin{align*}
&\E(\cL_{t+1} - \cL_{t}|\w_t) \leq -B\g_t^\top\left(\eta_\text{pub} \g_{t} + \eta_\text{priv} \E\frac{\g_{t, i}}{||\g_{t, i}|| + 1 }\right)\\
&+ \frac{L}{2}\Big(\eta_\text{pub}^2 \E \|\sum_j^{B} \g_{t, j}\|^2 + \eta_\text{priv}^2 \E \|\sum_i^{B} \frac{\g_{t, i}}{||\g_{t, i}|| + 1 }\|^2
\\
&+\eta_\text{priv}^2\sigma^2 d
+2\eta_\text{pub}\eta_\text{priv}B^2\E\g_{t,j}^\top\frac{\g_{t, i}}{||\g_{t, i}|| + 1 }\Big)
\end{align*}

It is not hard to see that
$$\E \|\sum_j^{B} \g_{t, j}\|^2=\|B \E\g_{t, j}\|^2+\text{\text{Var}}(\sum_j^{B} \g_{t, j})=B^2 \|\g_t\|^2+B\xi^2,$$
that we have
$\E\|\sum_i^{B} \frac{\g_{t, i}}{||\g_{t, i}|| + 1 }\|^2 \leq B^2$ from the AM-QM inequality and the fact that $\left\|\clip_i \g_{t, i}\right\| \le 1$, and that $\E(\g_{t,j}^\top\frac{\g_{t, i}}{||\g_{t, i}|| + 1 })=\g_t^\top \E(\frac{\g_{t, i}}{||\g_{t, i}|| + 1 })$ by the independence of $\g_{t,j}$ and $\g_{t,i}$. Therefore, we can write
\begin{align*}
\E(\cL_{t+1} - \cL_{t}|\w_t) 
&\leq -B\g_t^\top\left(\eta_\text{pub}\g_{t} + \eta_\text{priv} \E\frac{\g_{t, i}}{||\g_{t, i}|| + 1 }\right)\\
&+ \frac{L}{2}\Big( \eta_\text{pub}^2(B^2 \|\g_t\|^2+B\xi^2) + \eta_\text{priv}^2 B^2
\\
&+\eta_\text{priv}^2\sigma^2 d
+2\eta_\text{pub}\eta_\text{priv}B^2\g_t^\top \E(\frac{\g_{t, i}}{||\g_{t, i}|| + 1 })\Big)
\\
&\leq (\frac{L}{2}\eta_\text{pub}^2 B^2-\eta_\text{pub}B)\|\g_t\|^2+\frac{L}{2}\eta_\text{pub}^2 B\xi^2
\\
&+(L\eta_\text{pub}\eta_\text{priv}B^2-\eta_\text{priv}B) \g_{t}  \E(\frac{\g_{t, i}}{||\g_{t, i}|| + 1 })+ \frac{L}{2}\eta_\text{priv}^2 (B^2
+\sigma^2 d)
\end{align*}

Now we want to lower bound $\g_t^\top\E \left( \frac{\g_{t, i}}{||\g_{t, i}|| + 1} \right)$.
Based on the analysis of Theorem 6 in \cite{bu2022automatic}, we have for all $r>0$
$$
\g_t^\top \E\left( \frac{\g_{t, i}}{||\g_{t, i}|| + 1} \right) \geq \frac{1}{2} \mathcal{M}_r(||\g_t||),
$$
where $\mathcal{M}_r = \min_{0\leq c\leq 1} f(c, r; \frac{1}{||\g_t||}) \cdot (||\g_t|| - \xi/r)$.

By extending the expectation over randomness in the trajectory and summing over the iterations, we have
\begin{align}
    \cL_0
    &\geq \cL_0 - \E \cL_T = \sum_t\E(\cL_t - \cL_{t+1}) \\
    &\geq (\eta_\text{pub}B - \frac{L}{2}\eta_\text{pub}^2 B^2) \E(\sum_t||\g_t||^2) -\frac{L}{2}T\eta_\text{pub}^2 B \xi^2 \\
    &+ (\frac{\eta_\text{priv} B}{2}-\frac{L\eta_\text{priv} \eta_\text{pub} B^2}{2}) \E \sum_t \mathcal{M}_r(||\g_t||)- \frac{L}{2}T\eta_\text{priv}^2 \left( B^2 + \sigma^2 d\right) 
    \\
    &=(\eta_\text{pub}B - \frac{L}{2}\eta_\text{pub}^2 B^2)T \E(\frac{1}{T}\sum_t||\g_t||^2) -\frac{L}{2}T\eta_\text{pub}^2 B \xi^2 \\
    &+ (\frac{\eta_\text{priv} B}{2}-\frac{L\eta_\text{priv} \eta_\text{pub} B^2}{2})T\E \frac{1}{T} \sum_t \mathcal{M}_r(||\g_t||)- \frac{L}{2}T\eta_\text{priv}^2 \left( B^2 + \sigma^2 d\right) %
\end{align}
Up to this step, if we set $\eta_\text{priv}=0$ (i.e. not using the private gradient), we recover the public-data only convergence in Appendix D of \cite{bu2022automatic}; if we set $\eta_\text{pub}=0$ (i.e. not using the public gradient), we reduce to the private-data only convergence.

We apply the learning rate $\eta_\text{pub}=\frac{1}{BL\sqrt{T}}$ (same as \cite{bu2022automatic}; note that their gradient is mean reduction). Then we have,
\begin{align}
\cL_0 
    &\geq (\frac{\sqrt{T}-0.5}{L}) \E( \frac{1}{T}\sum_t ||\g_t||^2) - \frac{\xi^2}{2BL} \\
    &+ (\frac{\eta_\text{priv} B}{2}-\frac{\eta_\text{priv} B}{2\sqrt{T}})T \E \frac{1}{T} \sum_t \mathcal{M}_r(||\g_t||)- \frac{L}{2}T\eta_\text{priv}^2 \left( B^2 + \sigma^2 d\right) %
\\
    &\geq (\frac{\sqrt{T}}{2L}) \E( \frac{1}{T}\sum_t ||\g_t||^2) - \frac{\xi^2}{2BL}  \label{eq-forCompare}
 \\
    &+ (\frac{\eta_\text{priv} B}{2}-\frac{\eta_\text{priv} B}{2\sqrt{T}})T \E \frac{1}{T} \sum_t \mathcal{M}_r(||\g_t||)- \frac{L}{2}T\eta_\text{priv}^2 \left( B^2 + \sigma^2 d\right) \label{eq-forCompareEnd} %
\end{align}

For our asymptotic analysis, we consider large $T$ so that $\frac{\eta_\text{priv} B}{2}-\frac{\eta_\text{priv} B}{2\sqrt{T}}\sim \frac{\eta_\text{priv} B}{2}$.

Then applying $\eta_\text{priv} = \sqrt{\frac{2 \cL_0}{LT(1+\frac{\sigma^2d}{B^2})}} / B$, same as \cite{bu2022automatic},  we have
\begin{align}
\cL_0 
    &\gtrsim \frac{\sqrt{T}}{2L} \E(\frac{1}{T} \sum_t ||\g_t||^2) - \frac{\xi^2}{2BL}
    \\
    &+ \sqrt{\frac{T\cL_0}{2L(1+\frac{\sigma^2d}{B^2})}} \E \frac{1}{T} \sum_t \mathcal{M}_r(||\g_t||)- \cL_0
\end{align} %

For a brief notation, we denote $A_\text{pub} = \frac{1}{2L}$ and $A_\text{priv} = \sqrt{\frac{\cL_0}{2L(1+\frac{\sigma^2d}{B^2})}}$, thus we have
$$\frac{1}{\sqrt{T}}\left(2\cL_0+\frac{\xi^2}{2BL}\right)
\gtrsim A_\text{pub} \E(\frac{1}{T} \sum_t ||\g_t||^2)+ A_\text{priv}\E \frac{1}{T} \sum_t \mathcal{M}_r(||\g_t||)$$

$$\frac{1}{\sqrt{T}}\left(2\cL_0+\frac{\xi^2}{2BL}\right) \gtrsim \min_{0\leq t\leq T}\E\left(A_\text{pub} ||\g_t||^2+A_\text{priv} \mathcal{M}_r(||\g_t||)\right)$$

By Theorem 8 in \cite{bu2022automatic}, we have:
$$
\mathcal{M}_r(||\g_t||) = \left(\frac{1}{(r-1)||\g_t|| + 1} - \frac{1}{(r+1)||\g_t|| + 1}\right) \cdot (||\g_t||-\xi/r).
$$

Then,
\begin{align}
\frac{1}{\sqrt{T}}\left(2\cL_0+\frac{\xi^2}{2BL}\right)
&\gtrsim \min_{0\leq t\leq T}\E \cL_r(\|\g_t\|), \forall r>1
\end{align}

where
\begin{align}
\cL_r(\|\g\|)=A_\text{pub}\|\g\|^2+A_\text{priv}
\left(\frac{1}{(r-1)\|\g\|+1}-\frac{1}{(r+1)\|\g\|+1}\right) \cdot (\|\g\|-\xi/r)
\label{eq:fr explicit}
\end{align}

With Markov's inequality, we can complete the proof as follows: for any $a>0$,
$$\frac{1}{\sqrt{T}}\left(2\cL_0+\frac{\xi^2}{2BL}\right)
\gtrsim \min_{0\leq t\leq T}\E \cL_r(\|\g_t\|)\geq a\cdot\min_{0\leq t\leq T} \mathbb{P}(\cL_r(\|\g_t\|)>a)$$
Denote $\frac{1}{a\sqrt{T}}\left(2\cL_0+\frac{\xi^2}{2BL}\right)$ as $c$, then
\begin{equation}
\begin{aligned}
	{c} \gtrsim \min_{0\leq t\leq T} \mathbb{P}(\cL_r(\|\g_t\|)>a)
&=\min_{0\leq t\leq T} \mathbb{P}\left(\cL_r(\|\g_t\|)>\frac{1}{c\sqrt{T}}(2\cL_0+\frac{\xi^2}{2BL})\right) \\
&=\min_{0\leq t\leq T} \mathbb{P}\left(\|\g_t\|>\cL_r^{-1}\Big(\frac{1}{c\sqrt{T}}(2\cL_0+\frac{\xi^2}{2BL})\Big)\right)
\end{aligned}
\end{equation}

We can put in $r=1.5\xi/\|\g\|$ to roughly maximize $\cL_r$ since the above holds for any $r$:
\begin{align}
\cL_r(\|\g\|)\approx\frac{\|\g\|^2}{2L}+\sqrt{\frac{2\cL_0}{L(1+\frac{\sigma^2d}{B^2})}}\cdot \frac{\|\g\|^2}{3[(1.5\xi+1)^2 - ||\g||^2]}
\end{align}

\subsection{Proof of \Cref{cor:alpha}}\label{sec:proof cor:alpha}
\textit{Proof.}
In the proof of \Cref{thm:norm converge}, we have used $\eta_\text{pub}=\frac{1}{BL\sqrt{T}}$ and $\eta_\text{priv}= \sqrt{\frac{2 \cL_0}{LT(1+\frac{\sigma^2d}{B^2})}} / B$. 
Hence
\begin{align}
    \alpha=\frac{\frac{1}{BL\sqrt{T}}}{\frac{1}{BL\sqrt{T}}+\sqrt{\frac{2 \cL_0}{LT({B^2}+\sigma^2d)}}}=\frac{1}{1+B\sqrt{\frac{2 L\cL_0}{({B^2}+\sigma^2d)}}}.
\end{align}

Now we study the monotonicity: it is obvious that $\alpha$ is decreasing in $L,\cL_0$ and increasing in $\sigma, d$. To see that $\alpha$ is decreasing in $B$, it suffices to show 
\begin{align*}
 B\sqrt{\frac{2 L\cL_0}{({B^2}+\sigma^2d)}} \text{ is increasing}
\Longleftarrow  \frac{B^2}{({B^2}+\sigma^2d)} \text{ is increasing}
\end{align*}
which is obvious.

Also, from Equation (4), we have $\sigma \propto \frac{1}{n_{priv}}$.
Since $\alpha \propto \sigma$, we then have $\alpha \propto \frac{1}{n_{priv}}$.
Under $n=n_{pub} + n_{priv}$ and $r_{pub} = \frac{n_{pub}}{n_{pub} + n_{priv}}$, we have $\alpha \propto \frac{1}{n_{priv}} \propto n_{pub}$.

\subsection{Proof of \Cref{lam:function_fr}}\label{sec:proof_lam:function_fr}
\textit{Proof.}
To better understand the convergence, we simplify $f_r$ in \eqref{eq:fr explicit} via reasonable approximation that ignores unimportant terms: since \Cref{thm:norm converge} holds for any $r$, we substitute $r=2\xi/\|\g\|$ and derive the following approximation:
\begin{align*}
f_r(\|\g\|)&=\frac{\|\g\|^2}{2L}+\sqrt{\frac{\cL_0}{2L(1+\frac{\sigma^2d}{B^2})}}\frac{\|\g\|^2}{(2\xi+1)^2-\|\g\|^2}
\\
&\approx \left(\frac{1}{2L}+\sqrt{\frac{\cL_0 B^2}{2L\sigma^2 d}}\frac{1}{4\xi^2}\right)\|\g\|^2
\\
&\approx\left(\frac{1}{2L}+\sqrt{\frac{\cL_0  n_\textup{priv}^2\mu^2}{2LT d}}\frac{1}{4\xi^2}\right)\|\g\|^2,
\end{align*}
where the last row uses \Cref{cor:mu}. Note that by \eqref{eq:mu GDP}
\begin{align}
\sigma^2=1/\ln(\frac{n_\textup{priv}^2\mu^2}{TB^2} + 1)\approx \frac{B^2 T}{n_\textup{priv}^2 \mu^2}
\label{eq:sigma approx}
\end{align}
by the first order Taylor expansion. Therefore, we can approximate the gradient norm bound in \Cref{thm:norm converge} by
\begin{align*}
f_r^{-1}
&\approx \sqrt{\frac{\frac{1}{c\sqrt{T}}(2\cL_0+\frac{\xi^2}{2BL})}{\frac{1}{2L}+\sqrt{\frac{\cL_0  n_\textup{priv}^2\mu^2}{2LT d}}\frac{1}{4\xi^2}}}=\sqrt{\frac{\frac{1}{c\sqrt{T}}(2\cL_0 L+\frac{\xi^2}{2B})}{\frac{1}{2}+\sqrt{\frac{\cL_0 L n_\textup{priv}^2\mu^2}{2T d}}\frac{1}{4\xi^2}}}
\\
&=\frac{\sqrt{4\cL_0 L+\frac{\xi^2}{B}}}{c^{1/2}T^{1/4}}(1-x)+o(x)
\end{align*}
where we use $x=\sqrt{\frac{\cL_0 L n_\textup{priv}^2\mu^2}{2T d}}\frac{1}{4\xi^2}\to 0$ as $T\to\infty$.

\end{document}